\documentclass{article}
\PassOptionsToPackage{hyphens}{url}
\usepackage{iclr2026_conference,times}

\usepackage{amsmath,amssymb,amsthm}
\usepackage{graphicx}
\usepackage{placeins}
\usepackage{booktabs}
\usepackage{xcolor}
\usepackage{hyperref}
\usepackage{url}
\usepackage{array}
\usepackage{enumitem}
\graphicspath{{figures/}}

\newcommand{\yhat}{\hat{y}}
\newcommand{\yint}{\hat{y}_{\mathrm{int}}} 
\newcommand{\xstar}{x_*}            
\newcommand{\yctx}{\mathbf{y}}      
\newcommand{\ystd}{\tilde{y}}       
\newcommand{\ystdctx}{\tilde{\mathbf{y}}} 
\newcommand{\nctx}{n}               
\newcommand{\ai}{a_i}               
\newcommand{\effw}{a_i = \partial \yint / \partial \ystd_i} 
\newcommand{\permdelta}{\Delta_{\mathrm{perm}}}

\newcommand{\interdelta}{\Delta_{\mathrm{perm}}^{\perp}}   
\newcommand{\anisoH}{A_T}                                   
\newcommand{\crossx}{X_{\mathrm{sep}}}                      
\newcommand{\PT}{P_T}
\newcommand{\TabICL}{\textsc{TabICL-v2}}
\newcommand{\TabPFN}{\textsc{TabPFN}}
\newcommand{\TabPFNiii}{\textsc{TabPFN-v3}}
\newcommand{\TabPFNvtwosix}{\textsc{TabPFN-v2.6}}
\newcommand{\TabDPT}{\textsc{TabDPT}}
\newcommand{\TabFM}{\textsc{TabFM}}

\theoremstyle{plain}

\newtheorem{lemma}{Lemma}
\theoremstyle{definition}

\title{The Standardization Trap: \\ Certifying Joint Label Processing \\ in Tabular Foundation Models}

\author{Duong Nguyen, Nicolas Chesneau, and Milan Bhan \\
\texttt{\{first.last\}@ekimetrics.com}}

\iclrpreprintcopy
\hypersetup{
  pdftitle={The Standardization Trap: Certifying Joint Label Processing in Tabular Foundation Models},
  pdfauthor={Duong Nguyen, Nicolas Chesneau, Milan Bhan}
}

\begin{document}
\maketitle

\begin{abstract}
Linear regression and kernel smoothing offer tractable explanations of in-context learning: in both, the features determine the weight assigned to each context label. However, whether this fixed-weight account describes pretrained tabular foundation models (TFMs) remains unclear. Testing this account using derivatives runs into a standardization trap: public TFM packages standardize the labels before the model sees them, yet ordinary derivatives also reflect behavior outside the set of standardized labels, making a model appear nonlinear even when every prediction it makes agrees with a fixed-weight map. We propose two certificates that depend only on predictions at standardized labels and can reject two distinct explanations: fixed-weight prediction and sums of independent nonlinear label transformations. Across the five public TFMs that we evaluate, our certificates show that changing one context label alters how other labels influence the prediction, a behavior we call joint processing. We further find that joint processing emerges with training and that attention scores carry most of the measured interaction. Together, these findings motivate TFM explanations that account for how context labels change the influence of individual examples.
\end{abstract}

\section{Introduction}\label{sec:intro}

Influential studies have connected in-context learning to linear
regression and kernel smoothing
\citep{akyurek2023what,vonoswald2023transformers,zhang2024trained,collins2024incontext}.
In fixed-weight versions of these accounts, the features determine how the
examples are weighted, while the context labels supply the values being
combined. Predictions change when labels change, but the coefficients
do not. This gives a precise starting point for understanding adaptation
without parameter updates.

We ask how far this account extends to pretrained tabular foundation
models (TFMs). 
At inference, TFMs predict the label of a query using \emph{labeled} context examples while keeping their parameters fixed. How do TFMs use those context labels? Do the labels only supply the values being combined, or do they also change how the examples are weighted?

Consider predicting a house price from nearby sales. One strategy assigns a fixed
weight to each house using location and building attributes, then averages the observed
prices. Another also examines the prices themselves: an unusually high price
may be discounted when isolated but reinforced when nearby houses show the
same pattern. In the second strategy, one price changes the weight of 
other prices, even though the houses, their attributes, and the
query remain fixed. Do the
labels of a pretrained TFM only supply values that fixed weights combine, or
does one label also change the weight given to another?

To answer this question, we measure \emph{label sensitivity}: the first
derivative of the prediction with respect to one context label. 
For a model that assigns a fixed weight to a context observation, this sensitivity is the weight itself, so it does not move when the labels change. We therefore look at second derivatives and ask
whether the sensitivity to one label moves when another label changes.
Furthermore, a model could transform each
label separately, for example by limiting
extreme values, and then add the contributions. This way, the prediction might respond nonlinearly to label changes, while no label would change the weight of another. We therefore distinguish
fixed affine prediction, independent nonlinear label transformations, and
interaction across examples. That distinction allows us to precisely ask what kind of adaptation
a model has learned, rather than treating every nonlinear response as the
same phenomenon.

Measuring such label interactions with derivatives runs into a surprising trap: tests based on derivative analysis can suggest that labels interact even when every prediction that the model makes combines the labels with fixed weights. Label standardization restricts the labels entering the network to vectors with a fixed mean and scale. The raw labels
never reach the network directly, yet ordinary derivatives also describe
behavior away from the set of standardized labels. For example, adding to a fixed-weight map a component that is zero on this set leaves every prediction unchanged but changes the derivatives (Section \ref{subsec:followon}). We could therefore
mistake behavior that never affects the prediction for evidence of
adaptation.

Our contributions are as follows:
\begin{enumerate}[noitemsep,topsep=0pt,leftmargin=*]
\item \textbf{Certificates that distinguish three forms of label processing}:
We derive two certificates that account for standardization and distinguish three forms of label processing: (1) fixed-weight prediction, in which the weights do not depend on the labels, (2) sums of independent nonlinear label transformations, in which a label changes only its own influence, and (3) joint processing, in which one label changes the influence of the others. A nonzero value of the first certificate excludes (1), and a nonzero value of the second excludes both (1) and (2), which establishes (3). 

\item \textbf{Evidence of joint processing and its association with training}: 
Using the proposed certificates, we show evidence of joint processing in the internal readouts of all tested models. Randomly initialized copies of the same architectures also respond nonlinearly to labels, but they show far less interaction across examples, which indicates that joint processing appears with training rather than coming from the architecture alone.

\item \textbf{Identification of the derivative paths carrying the measured interaction}: We show that holding the attention scores fixed during differentiation removes most of the
measured local curvature while leaving the predictions unchanged, so most of the measured interaction passes through the attention scores.

\end{enumerate}

Our paper is organized as follows. Section~\ref{sec:related} reviews related work on explanations of how in-context learning works. Section~\ref{sec:method} introduces the standardization trap and derives our two certificates. Section~\ref{sec:results} applies these certificates to five public TFMs and examines the roles of training and attention mechanisms in joint label processing. Finally, Section~\ref{sec:results-positive} discusses the implications and limitations of our findings.

\section{Related work}\label{sec:related}

\paragraph{Fixed-weight accounts of in-context learning.}
Several prior works have shown that a transformer can carry out linear regression, or steps of gradient descent, on the context examples
\citep{vonoswald2023transformers,akyurek2023what,ahn2023transformers,zhang2024trained}, 
and attention can be interpreted as a kernel operation
\citep{tsai2019transformer,collins2024incontext}.  In the simple versions of these explanations, the prediction is a weighted sum of the context labels. The features then characterize the weight of each example. The labels only supply the values to be combined (i.e., changing a label never changes a weight). We test whether pretrained TFMs process context labels the same way. With the features fixed, can one set of weights explain the predictions for every set of labels?

Writing a model in kernel form does not settle the question, because the kernel itself may
already depend on the labels. For example, \citet{kernelicl} fine-tune TabICL \citep{tabicl} so that its final prediction is a kernel average, but the inputs of this kernel are embeddings that the model has computed from the features and the
labels together. 
Any adaptation to the labels has then happened before the kernel is applied. The whole computation therefore has to be examined, including how representations are built and how parameters such as a bandwidth are chosen.

\paragraph{Learning a task from its labels.}
In other accounts, the labels do more than supply values. They also help decide which map is
applied. Prior-data fitted networks are trained to approximate a Bayesian predictor. Such a predictor updates its belief about the task from the observed labels \citep{muller2022transformers}.
Bayesian accounts of in-context learning make the same point: the labeled examples reveal
which task is being solved, and the prediction follows from that task
\citep{xie2022explanation,panwar2024incontext}. A transformer can also select among several
algorithms according to how well each one predicts examples held out from its fit
\citep{bai2023transformers}. In all three accounts, changing the labels can change the
inferred task or the selected algorithm, and with it the weight given to every label. Our
measurements agree with this picture. \citet{nagler2023statistical} addresses a different
question, why the bias of a PFN need not vanish as the context grows unless the model focuses on the examples near the query, whereas we
study how a model responds when the labels change at a fixed context size.

\paragraph{From internal computation to prediction.}
Another line of work studies the internal computation of TFMs. \citet{gupta2026lookingglass} show that the hidden layers of TabPFN encode intermediate quantities of this computation, such as partial products of an arithmetic target, and \citet{ye2025} find that TabPFN~v2 \citep{tabpfn} can serve as a feature extractor whose features separate the classes well. Predictions are refined gradually across layers,
with so much overlap that one layer applied six times matches six distinct layers in a small model
\citep{balef2026onelayer}. Ablation experiments find that one attention head dominates the
others at its peak layer \citep{gupta2026wherecomputation}. Model families also read the
context labels in different ways, from an attention-weighted vote over the labels to a
class-mean readout \citep{tfmmech2026}. Under noisy labels and irrelevant features, TabPFN's
attention stays on the informative features \citep{hu2026noise}.
Benchmarks measure how well the models predict
\citep{grinsztajn2022why,erickson2025tabarena,purucker2026beyondiid}.
For language models,
some studies ask how much a prediction depends on the labels of the demonstrations
\citep{min2022rethinking,kossen2024incontext}. Other studies connect in-context learning to gradient
descent, induction heads, and task vectors
\citep{dai2023gpt,olsson2022induction,hendel2023taskvectors,todd2024functionvectors}, or trace
a prediction back to individual demonstrations \citep{detail2024}.
These studies leave one research question open: does one label change how another label is used? We address this question in the following section. 

\section{Measuring how models use labels}\label{sec:method}

This section presents how we address the research question introduced above. 
The objective is to measure how a TFM uses its context labels, and in particular whether changing one label changes how other labels influence the prediction. Derivatives are the natural tool, but using them directly runs into a trap: label standardization confines the labels reaching the network to a proper subset of the label space, while derivatives in ambient coordinates also describe behavior off that subset. A model that coincides with a fixed affine map on every admissible label vector can therefore exhibit nonzero second derivatives. We develop tests that remove this ambiguity, distinguishing joint processing from fixed affine prediction and from independent nonlinear transformations of individual labels.

\subsection{From a label's value to its influence}\label{subsec:hyp}

Let $i\in\{1,\ldots,n\}$ index the context examples. Each example has a
feature vector $x_i\in\mathbb R^d$ and a label $y_i$. The context feature matrix and
corresponding label vector are
$X={[x_1\cdots\,x_n]}^\top\in\mathbb R^{n\times d}$ and
$\yctx={(y_1,\ldots,y_n)}^\top\in\mathbb R^n$, respectively.
For a query with feature vector $x_*$, write the prediction as $\yhat=f(x_*,X,\yctx)$.
Note that the network itself does not receive the raw labels
$\yctx$. A TFM package first standardizes $\yctx$ to $\ystdctx$ in a wrapper, with components
$\ystd_i=(y_i-\bar y)/s_y$, where $\bar y$ is the context-label mean and
$s_y$ is the standardization scale. The network computes the internal
prediction $\yint=g(\ystdctx)$, and the output conversion returns $\yhat$
in the original label units. This distinction determines what we measure:
the response of the internal map to standardized labels, with one estimator
and a fixed output conversion. 

We describe this response through the \emph{label sensitivity}
$a_i(\ystdctx)=\partial g(\ystdctx)/\partial \ystd_i$, which measures how strongly
label $i$ affects the prediction. Several existing
interpretations of in-context learning describe the prediction as affine in the labels:
\begin{equation}
\yint=b+\sum_{i=1}^{\nctx}\ai(x_*,X)\,\ystd_i,
\label{eq:fixedweight}
\end{equation}
where the offset $b$ and the weights $\ai(x_*,X)$ depend on the features and the query but not on the labels. For example, the
kernel-smoothing interpretation of \citet{collins2024incontext} takes a weighted
average of the labels, with nonnegative weights that sum to one. The converged
single-layer linear-attention predictor analyzed by \citet{zhang2024trained}
allows weights of either sign. In these accounts, the sensitivities are the
weights themselves, and stay constant when the context labels change. We call
such a map a \emph{fixed-weight} map.

Other interpretations let the labels change the weights themselves, so that $\ai=\ai(x_*,X,\ystdctx)$. A map whose sensitivities change with the labels can still be of two kinds. It can transform \emph{one label at a time} and add the results, so
that a label changes only its own sensitivity, or the labels can interact, which
we call \emph{joint processing}.

To test whether a model uses a fixed-weight map, one can perturb the labels and
compare the sensitivity vectors before and after the perturbation. A fixed-weight
map returns the same vector twice, so a large change would seem to show that the
weights are not fixed. In the next section, we show that this test falls into a trap, and so does
the second-derivative test that extends it. We then present protocols that avoid
this trap and certify how a model uses its context labels.

\begin{figure}[t]\centering
\includegraphics[width=0.74\linewidth]{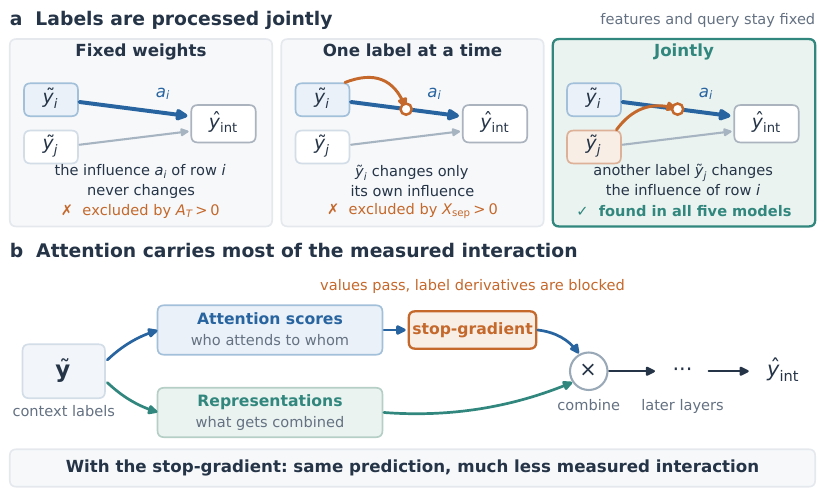}
\caption{\textbf{Labels are processed jointly, and attention carries most of the measured
interaction.}
\textbf{a}: three ways a model could use context labels while the features and the query stay
fixed. The blue arrow is the influence $a_i$ of row $i$ on the internal prediction
$\yint$. With fixed weights, this influence never changes. If labels are handled one at a time,
a label can change only its own influence. In joint processing, another label $\tilde y_j$
changes how much row $i$ matters, which is what all five models show
(\S\ref{sec:results-hessian}).
\textbf{b}: inside an attention step, labels can act through the attention scores and
through the representations that those scores combine. A stop-gradient, applied at every
attention step, lets the scores pass unchanged but stops labels from influencing them
when derivatives are taken. The prediction stays the same, while most of the measured
interaction disappears (\S\ref{sec:results-origin}). The stop-gradient shows which derivative paths
carry the interaction in the computation as implemented, not that these paths are necessary.}
\label{fig:jointattention}
\end{figure}

\subsection{The standardization trap}\label{subsec:followon}

\begin{figure}[t]\centering
\includegraphics[width=0.72\linewidth]{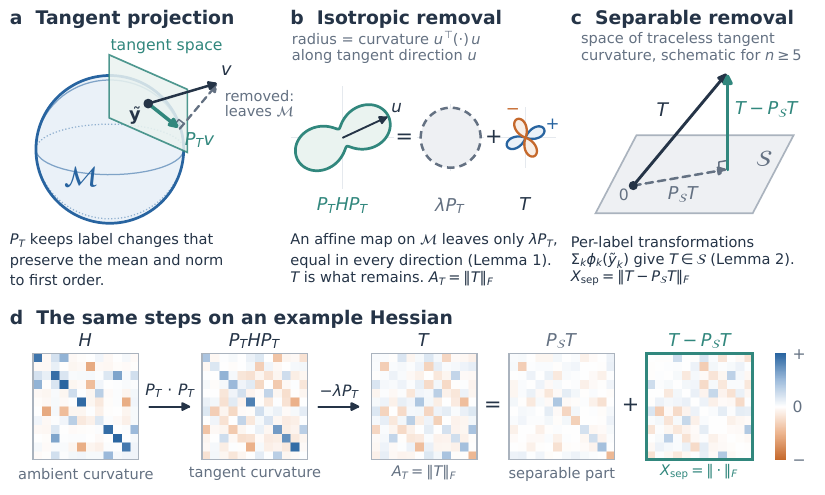}
\caption{\textbf{The standardization trap and the two certificates} (\S\ref{subsec:followon}).
\textbf{a}: the network only ever sees labels on the standardization manifold $\mathcal M$, but
derivatives also react to label changes that would leave it. For a label perturbation
$v\in\mathbb R^n$, the projector $\PT$ keeps the part that slides along $\mathcal M$, so that
a small step keeps the label mean and changes the label norm only negligibly.
\textbf{b}: each shape shows how strongly the prediction curves along every direction $u$ within
the tangent space, drawn as the distance from the center. This curvature splits into a part that
is equal in every direction, $\lambda\PT$, and the remainder $T$. A function such as $g_a$ that
is affine on $\mathcal M$ can produce only the first part (Lemma~\ref{lem:isotropy}), so
$\anisoH=\lVert T\rVert_F$ measures what such a map cannot explain. The equal part is enlarged
for visibility.
\textbf{c}: $\mathcal S$ contains every curvature pattern that arises when each label is
transformed separately, as in $g_{a,\phi}$. $P_{\mathcal S}T$ is the closest such pattern to
$T$, and the length of what is left over is $\crossx$ (Lemma~\ref{lem:separable}). 
\textbf{d}: the same steps on a constructed Hessian. It sums terms that act only
off $\mathcal M$, a separate effect for each label, and couplings between pairs of labels. Only
the pair couplings reach $T-P_{\mathcal S}T$.}
\label{fig:certgeometry}
\end{figure}

A natural way to test for fixed weights is to differentiate the sensitivities once
more. The Hessian measures how the sensitivity to label $i$ changes when label $j$
changes:
\[
H_{ij}(\ystdctx)=\frac{\partial a_i}{\partial \ystd_j}
=\frac{\partial^2 g}{\partial \ystd_i\,\partial \ystd_j}.
\]
A fixed-weight map has $H=0$, and it is tempting to conclude in turn that a
nonzero $H$ means that the weights are not fixed. This conclusion falls into \emph{the standardization
trap}. The rest of this section explains the trap on an example and how to avoid it.

The trap comes from the standardization step (Section~\ref{subsec:hyp}), which
does more than change the label units. Because it fixes the mean
and the squared norm of the labels, it confines the inputs of the network to the
\emph{standardization manifold}
\[
 \mathcal M=\{\ystdctx\in\mathbb R^n:\mathbf 1^\top \ystdctx=0,\ \lVert \ystdctx\rVert^2=r^2\},
\]
where $\mathbf 1$ is the all-ones vector and $r$ is the norm that the scale $s_y$
imposes, with $r^2=n$ for the population standard deviation. This manifold is a
sphere inside the subspace of mean-zero vectors. Derivatives, however, also
describe how $g$ behaves when the labels leave this sphere, which never happens
in use.

As a demonstration, consider a fixed-weight map as in
Eq.~\eqref{eq:fixedweight}, with offset $b$ and weights $a\in\mathbb R^n$, and add
a term that vanishes on $\mathcal M$:
\begin{equation}
 g_a(\ystdctx)=b+a^\top\ystdctx+q\,\bigl(\lVert\ystdctx\rVert^2-r^2\bigr),\qquad q\neq0.
 \label{eq:ga}
\end{equation}
On every valid input, the added term is zero. Hence, $g_a$ and the fixed-weight map
$b+a^\top\ystdctx$ make exactly the same predictions, and no experiment that feeds
labels to the model can tell them apart. Their derivatives nevertheless differ.
The sensitivities of $g_a$ are $a+2q\ystdctx$, so they change whenever the labels
are permuted, and its Hessian is $2qI$, which can be arbitrarily large.
\emph{Neither effect changes a single valid prediction.} Naive tests would
therefore report that $g_a$ adapts its weights to the labels, although it applies
fixed weights wherever it is ever evaluated. Nothing prevents a trained network
from containing such terms, because training never evaluates the network away from
$\mathcal M$ and so never constrains how it extends there.

For any map $g$, a valid test must therefore depend only on the values that $g$ takes on
$\mathcal M$.
The first step in building such a test is to keep only the label changes that stay on $\mathcal M$.
A change $v\in\mathbb R^n$ of the labels keeps their mean when its entries sum to zero, and for small steps it keeps their norm when it is perpendicular to the
current label vector $\ystdctx$. The changes that satisfy both conditions are the
\emph{tangent directions} at $\ystdctx$: they slide along the sphere instead of
leaving it (Figure~\ref{fig:certgeometry}a). The matrix $P_T=I-\frac{\mathbf 1\mathbf 1^\top}{n}-\frac{\ystdctx\ystdctx^\top}{r^2}
$
turns any change into a tangent one. 
The Hessian $H_{\ystdctx}=\nabla_{\ystdctx}^2g(\ystdctx)$ involves two
label changes at once, so we apply $\PT$ on both sides.
This gives the \emph{tangent curvature} $\PT H_{\ystdctx}\PT$, which records how the
sensitivities change when the labels move along tangent directions only.

This correction is not enough. For example, going back to our example with $g_a$, the Hessian $2qI$ gives the tangent curvature $2q\PT$ (since $P_T$ is idempotent), which is not zero, so the test still fires. The reason is that $\mathcal M$ is curved, whereas a tangent direction is a
straight line that touches $\mathcal M$ at one specific point only. A step of length $\delta$
along a unit tangent direction $v$ therefore leaves the sphere slightly, since
$\lVert\ystdctx+\delta v\rVert^2=r^2+\delta^2$. The gap $\delta^2$ is too small to affect the
slope of a function, but second derivatives measure exactly the terms in
$\delta^2$, and the added term of $g_a$ turns this gap into the value $q\delta^2$.

The gap $\delta^2$ does not depend on the direction $v$ of the step. The tangent
curvature $2q\PT$ that it produces for $g_a$ therefore has the same value $2q$
in every tangent direction. We call a curvature that is equal in all tangent
directions \emph{isotropic}. Lemma~\ref{lem:isotropy} shows that this form is not special to
$g_a$: every function that agrees with a fixed-weight map on $\mathcal M$ has an
isotropic tangent curvature, whatever it does away from $\mathcal M$.

\begin{lemma}[Affine predictions on the manifold]\label{lem:isotropy}
Let $g$ be any smooth function on a neighborhood of $\mathcal M$ that agrees on
$\mathcal M$ with an affine function of $\ystdctx$. Then at every
$\ystdctx\in\mathcal M$,
\[
 \PT H_{\ystdctx}\PT=\lambda(\ystdctx)\,\PT
\]
for some scalar $\lambda(\ystdctx)$.
\end{lemma}

We provide the proof of Lemma~\ref{lem:isotropy} in Appendix~\ref{app:manifold}. Every function covered by the lemma is built like $g_a$: a fixed-weight
map plus terms that vanish on $\mathcal M$. By Lemma~\ref{lem:isotropy}, such a map can leave
only isotropic curvature. We therefore define the operator $T$ as the tangent curvature minus its isotropic part, and we measure its
size $\anisoH$:
\begin{equation}
 T=\PT H_{\ystdctx}\PT-
 \frac{\operatorname{tr}(\PT H_{\ystdctx}\PT)}{n-2}\PT,
 \qquad \anisoH=\lVert T\rVert_F,
 \label{eq:aniso}
\end{equation}
where $\lVert\cdot\rVert_F$ is the Frobenius norm. The trace of the tangent curvature is the sum of the curvatures along $n-2$ perpendicular tangent directions, so dividing it by $n-2$ gives the average curvature over the tangent directions (Figure~\ref{fig:certgeometry}b), and $T$ keeps only the departure from this average. 
For $g_a$, whose Hessian is $H_{\ystdctx}=2qI$, both
quantities in Eq.~\eqref{eq:aniso} can be computed by hand:
\begin{align*}
\PT H_{\ystdctx}\PT&=\PT(2qI)\PT=2q\,\PT\PT=2q\,\PT,\\
\operatorname{tr}(\PT H_{\ystdctx}\PT)&=2q\operatorname{tr}(\PT)=2q\,(n-2).
\end{align*}
The first line uses $\PT\PT=\PT$, since projecting twice changes nothing, and the
second line uses $\operatorname{tr}(\PT)=n-2$, since $\PT$ removes two directions
from the $n$-dimensional identity. Substituting both lines into
Eq.~\eqref{eq:aniso} gives
$T=2q\,\PT-\tfrac{2q\,(n-2)}{n-2}\,\PT=0$,
so the corrected test stays silent on $g_a$, as it should. 
The value of $q$ plays no role in this cancellation, so the same argument gives $T=0$ for every isotropic curvature $\lambda\PT$, regardless of the value of $\lambda$.
More generally, $T$ depends only on the predictions on $\mathcal M$, regardless of how $g$ extends away from the manifold. \emph{A nonzero $\anisoH$ therefore excludes fixed-weight prediction on valid labels}.

A nonzero $\anisoH$ shows that $g$ is nonlinear on $\mathcal M$, but nonlinearity alone is not enough to establish joint processing. A model can
transform one label at a time and add the results, as in
\begin{equation}
 g_{a,\phi}(\ystdctx)=g_a(\ystdctx)+\sum_{k=1}^{n}\phi_k(\ystd_k),
 \label{eq:gaphi}
\end{equation}
where each smooth function $\phi_k$ acts on a single label, for example by damping
extreme values. For most choices of $\phi_k$, this map is not affine on $\mathcal M$, and it has
$\anisoH>0$. Yet no label changes the influence of another one, since the added
term contributes only the diagonal matrix
$D=\operatorname{diag}\bigl(\phi_k''(\ystd_k)\bigr)$ to the Hessian. We cannot detect
this case by looking for zero off-diagonal entries, because the projection mixes the
coordinates: on $\mathcal M$ no label can move alone, since the other labels must
compensate to keep the mean and the norm fixed, and $\PT D\PT$ is no longer diagonal.
We test instead whether the corrected curvature $T$ could have come from
\emph{some} diagonal matrix. Let
$\mathcal S=\{\PT D\PT-\tfrac{\operatorname{tr}(\PT D\PT)}{n-2}\PT:
D\ \text{diagonal}\}$
be the linear space of corrected curvatures that one-label-at-a-time maps, such as
$g_{a,\phi}$, can
produce, and let $P_{\mathcal S}$ denote the orthogonal projection onto it. We
measure what remains outside this space:
\begin{equation}
 \crossx=\lVert T-P_{\mathcal S}T\rVert_F.
 \label{eq:crossx}
\end{equation}
For $g_{a,\phi}$ we have $T\in\mathcal S$ and hence $\crossx=0$, whatever the
functions $\phi_k$ and the coefficient $q$. Figure~\ref{fig:certgeometry}
illustrates both projections, and Lemma~\ref{lem:separable} states the general
result.

\begin{lemma}[Certificate of joint label processing]\label{lem:separable}
Let $g$ be smooth on a neighborhood of $\mathcal M$ and agree on $\mathcal M$
with an affine-plus-separable map
$b+a^\top\ystdctx+\sum_k\phi_k(\ystd_k)$.
Then $T\in\mathcal S$ and $\crossx=0$ at every point of $\mathcal M$.
The statistic $\crossx$ is extension-invariant: any two smooth maps that
agree on $\mathcal M$ give the same value there.
\end{lemma}

\emph{A nonzero $\crossx$ therefore certifies that one label changes the influence of
another}. Both tests are one-sided: a zero result alone does not prove membership
in the corresponding class. Table~\ref{tab:certsummary} follows the two examples and a joint-processing map through each correction. We provide in Appendix~\ref{app:manifold} the
proofs, the computation of $P_{\mathcal S}$, and numerical checks of this pattern.


\begin{table}[htb]
\centering\small
\caption{\textbf{From the Hessian to the two certificates.} Each row
applies one more correction to the Hessian $H$ of the map. The last three columns give the
result for $g_a$, which is a fixed-weight map on $\mathcal M$ (Eq.~\ref{eq:ga}), for the
one-label-at-a-time map $g_{a,\phi}$ (Eq.~\ref{eq:gaphi}, with
$D=\operatorname{diag}(\phi_k''(\ystd_k))$), and for a map in which labels interact. Only $\anisoH$ vanishes for every fixed-weight map, and
only $\crossx$ also vanishes for every one-label-at-a-time map.}
\label{tab:certsummary}
\setlength{\tabcolsep}{4pt}
\begin{tabular}{llccc}
\toprule
Quantity & What it keeps & $g_a$ & $g_{a,\phi}$ & Joint map \\
\midrule
Hessian $H$ & every label change & $2qI$ & $2qI+D$ & $\neq0$ \\
Tangent curvature $\PT H\PT$ & changes that stay on $\mathcal M$ & $2q\PT$ & $2q\PT+\PT D\PT$ & $\neq0$ \\
$T$, of size $\anisoH$ & minus the isotropic part & $0$ & in $\mathcal S$, $\neq0$ & $\neq0$ \\
$T-P_{\mathcal S}T$, of size $\crossx$ & minus the one-label-at-a-time part & $0$ & $0$ & $\neq0$ \\
\bottomrule
\end{tabular}
\end{table}

The absolute values of $\anisoH$ and $\crossx$ themselves are hard to interpret, because they
depend on the scale of the prediction and of its sensitivities, which differs
between models and datasets. We therefore report two shares:
\begin{equation}
 \bar{A}_T=\frac{\anisoH^2}{\lVert\PT H_{\ystdctx}\PT\rVert_F^2},
 \qquad
 \bar{X}_{\mathrm{sep}}=\frac{\crossx^2}{\anisoH^2}.
 \label{eq:shares}
\end{equation}
Both lie between $0$ and $1$, because each correction removes a part that is
orthogonal to the part it keeps, and we set a share to $0$ when its denominator
vanishes. $\bar{X}_{\mathrm{sep}}$ is the share of the
corrected curvature that no one-label-at-a-time map can produce, and like
$\crossx$ it depends only on the predictions on $\mathcal M$. $\bar{A}_T$ is
different: its denominator still contains the isotropic part of the
tangent curvature, which depends on how $g$ behaves away from $\mathcal M$.
Apart from vanishing exactly when $\anisoH$ does, $\bar{A}_T$ serves another purpose: $1-\bar{A}_T$ is the fraction of the computed
tangent curvature that the isotropic correction subtracts, so it shows how much
this correction changes the measurement in the tested networks. A fixed-weight
map has $\bar{A}_T=0$, and a one-label-at-a-time map has
$\bar{X}_{\mathrm{sep}}=0$.


\section{Experiments and results}\label{sec:results}

Our experiments answer two questions about pretrained TFMs: (1) do the models
process their context labels jointly, which is the question that the two
certificates were built to settle (Section~\ref{sec:results-hessian}), and
(2) where does the interaction come from, both whether it is already present
before training and which paths inside the network carry it
(Section~\ref{sec:results-origin}).
Section~\ref{subsec:arms} first presents the setup that all experiments share.

\subsection{Experimental setup}\label{subsec:arms}

\paragraph{Models.}
We evaluate five public TFMs: \TabICL{} \citep{tabiclv2}, \TabPFNiii{} \citep{tabpfn3},
\TabPFNvtwosix{} \citep{tabpfn25}, \TabFM{} 1.0 \citep{tabfm}, and \TabDPT{} v1.2
\citep{tabdpt,tabdptturbo}. They differ in architecture, attention schedule, readout, and
pretraining data (Appendix~\ref{app:arms}). The first four are pretrained on
synthetic tasks and have related designs: \TabICL{}, \TabPFNiii{} and \TabFM{}
build a representation of each row with attention across features and rows and
then predict in a final in-context stage, whereas \TabPFNvtwosix{} alternates
attention across features and rows in every layer. \TabDPT{} is pretrained on real tables, which makes it the most
independent of the five.

\paragraph{Tasks and Protocol.}
We compute the two certificates on four regression datasets with $256$ context
rows: one synthetic dataset and the OpenML datasets \texttt{kin8nm}, \texttt{pol},
and \texttt{superconduct}. The synthetic dataset is a smooth regression task on
$16$ continuous features that we generate ourselves (Appendix~\ref{app:generator}),
so that it serves as a control that real tables cannot provide: we can redraw its
label noise from the true model, no model can have seen it during pretraining, and
it has no outliers or discrete features that could explain what we measure.
We call one model on one dataset a \emph{setting}, which
gives $20$ settings. In each setting, we repeat the measurement on three randomly sampled contexts, with four queries each. Appendix Table~\ref{tab:cells} lists
the datasets, context sizes, and repetitions of every experiment in the paper.

\paragraph{Implementation Details.} Since we run our analysis based on TFM responses, we take
derivatives with respect to the standardized labels and keep the conversion back to the
original label units fixed (Section~\ref{subsec:hyp}). The differentiated prediction
comes from a single estimator inside each package, whereas the packages' default
predictions combine several estimators, which we test separately
(Section~\ref{sec:results-hessian}). Second derivatives need an attention implementation
that can be differentiated twice, which the fused kernels used in deployment are not, so
we switch to standard (eager) attention for these computations (Appendix~\ref{app:hvp}). 

\subsection{To what extent do TFMs process labels jointly?}\label{sec:results-hessian}

We first determine which of the three descriptions of
Section~\ref{subsec:hyp} fits each TFM: fixed weights, one label at a time, or
joint processing (Figure~\ref{fig:jointattention}a). To this end, we
compute the exact Hessian of the internal prediction in each of the $20$ settings
and apply the two certificates, so that each description is tested by a statistic
that is zero whenever the description holds. Table~\ref{tab:main} reports the
results.

\begin{table}[ht]
\centering\small
\caption{\textbf{All five models process labels jointly, and most of the interaction passes
through attention.} The shares (Eq.~\eqref{eq:shares}) and the relative fall of
$\anisoH$ and $\crossx$ when derivatives cannot pass through the attention scores
(Section~\ref{sec:results-origin}), in percent, each averaged over the four datasets.
A fixed-weight map would give $\bar{A}_T=0$, and a one-label-at-a-time map would give
$\bar{X}_{\mathrm{sep}}=0$. Appendix Table~\ref{tab:crossx} gives every setting.}
\label{tab:main}
\begin{tabular}{lcccc}
\toprule
 & \multicolumn{2}{c}{Share (\%)} & \multicolumn{2}{c}{Attention drop (\%)} \\
\cmidrule(lr){2-3}\cmidrule(lr){4-5}
Model & $\bar{A}_T$ & $\bar{X}_{\mathrm{sep}}$ & $\anisoH$ & $\crossx$ \\
\midrule
\TabICL{} & $98.7$ & $63.1$ & $81.6$ & $82.0$ \\
\TabPFNiii{} & $98.5$ & $70.8$ & $85.9$ & $90.0$ \\
\TabPFNvtwosix{} & $97.7$ & $73.4$ & $90.9$ & $95.3$ \\
\TabFM{} & $99.2$ & $67.5$ & $78.2$ & $82.1$ \\
\TabDPT{} & $98.3$ & $40.4$ & $68.0$ & $75.3$ \\
\midrule
All models & $98.5$ & $63.0$ & $80.9$ & $85.0$ \\
\bottomrule
\end{tabular}
\end{table}

Fixed weights do not explain the predictions of any model, since $\bar{A}_T$ is
nonzero in every setting (Appendix Table~\ref{tab:crossx}).
Its average of $98.5\%$ in Table~\ref{tab:main} also shows that the isotropic correction
subtracts only $1.5\%$ of the computed tangent curvature, so for these models
the correction changes the measurement little, although the test would not be
valid without it.
Transforming one label at a time does not explain the predictions either. Such a
map would give $\bar{X}_{\mathrm{sep}}=0$, whereas $\bar{X}_{\mathrm{sep}}$ averages
$63\%$ over the $20$ settings and lies between $23\%$ and $92\%$ in every one of
them (Appendix Table~\ref{tab:crossx}). 

The package default inference procedures of the tested models combine several
estimators. We tested those settings and found that the combined predictions are
not affine in the labels either (Appendix~\ref{app:robustness}). 

\subsection{Where do the interactions come from?}\label{sec:results-origin}

The certificates analyzed above establish that labels interact inside the models, but they do
not say where this interaction comes from. We first test whether the architecture
alone produces the interaction or whether it appears with training, and we then
identify which paths inside the network carry it.

\paragraph{Training changes the interaction among labels.}\label{sec:results-training}
We want to know whether the interaction comes with the architecture or appears
with training. We therefore compare each trained model with $12$ randomly
initialized copies of the same architecture on two of the four datasets. A randomly initialized network is
already a nonlinear function of its labels, so it departs from fixed weights
before training: $\anisoH$ is nonzero for both the trained models and the random copies and therefore cannot tell them apart. 
The normalized certificate $\bar{X}_{\mathrm{sep}}$ does separate trained from
untrained networks. On the two datasets of this comparison, every query of a trained
model gives at least $29\%$, whereas the random copies stay at most $3.4\%$ (Appendix~\ref{app:followups}). 



\paragraph{Attention carries most of the measured interaction.}\label{sec:results-routing}
Attention combines representations with weights that it computes from attention
scores, and labels can change both the representations being combined and the
scores that weight them. To separate these contributions, we block derivatives
through the attention scores while keeping their values unchanged in the forward
computation (Figure~\ref{fig:jointattention}b), so that the model makes exactly
the same prediction while its derivatives retain only the paths that do not pass
through these scores. Blocking these derivative paths removes most of the measured interaction
(columns ``Attention drop'' in Table~\ref{tab:main}). The cross-example residual $\crossx$
falls by $85\%$ on average, from $75\%$ for \TabDPT{} to $95\%$ for
\TabPFNvtwosix{}, and the corrected curvature norm $\anisoH$ falls by $81\%$ on
average. These percentages describe reductions in norms, whereas the comparison with
random copies uses shares of squared curvature, and they neither measure
accuracy loss nor establish that these paths are necessary for good
predictions. Appendix~\ref{app:mechanism} compares the curvature operators directly and
describes a forward intervention. These results add to the ongoing debate on whether attention weights explain the predictions of a model~\citep{wiegreffe2019attention, jain2019attention, bibal2022attention, bhan2023evaluating}.

\section{Discussion and conclusion}\label{sec:results-positive}

In the evaluated models, context labels can change how other labels
affect the prediction. The packages standardize the labels before the
networks see them, and ordinary derivatives also describe how the network
would behave on labels that are not standardized, so they can report
nonlinearity even for models that apply fixed weights to every valid
input. Our two certificates measure behaviors on standardized labels only. The first certificate $\anisoH$ is
zero for every fixed-weight map, and the second certificate $\crossx$ is
zero for every map that transforms one label at a time. Both are nonzero in all five models, 
so no model is explained by either fixed weights or transforming its labels one at a time.
Of the curvature that departs from fixed weights, $23\%$ to $92\%$ comes
from interaction that no one-label-at-a-time map can produce. This
interaction is much weaker in randomly initialized copies of the same
architectures, which indicates that it comes with training rather than from
the architecture alone. Blocking derivatives through the attention scores
removes most of it while leaving the predictions unchanged, so in the networks
as implemented most of the interaction passes through these scores.

Our findings guide explanations of how TFMs use their
context labels. They exclude the fixed-weight account, in which the features alone set the weight of each label, because in every evaluated model a change in one label also changes the weights of the others. Yet they do not identify the computation behind this interaction, since several computations make labels interact: (1) the model may use all labels to infer which task generated the data, so that changing any label changes the inferred task and with it the weight of every label, (2) a local regression may reweight the neighbors of the query by their labels, and (3) the representations of the examples may themselves depend on the labels. Our certificates detect the interaction but cannot tell these computations apart. The fixed-weight account also remains a useful approximation even though it fails as an exact description: a fixed-weight map fitted to a model's predictions explains a median of 72\% of the variation in these predictions when the labels are redrawn from the same task (Appendix~\ref{app:tuning}). An explanation should therefore state for which labels it holds and whether it reproduces only the predictions or also how one label changes the influence of the others.


The remaining question is whether these label interactions improve
prediction, since our derivative tests establish their presence but not
their necessity. Blocking the attention scores does not answer it
either, because the predictions stay the same.
Resolving it requires an intervention that removes the interaction from the forward computation and a measurement of the resulting change in predictive accuracy. 

\clearpage
\FloatBarrier
\typeout{MANUSCRIPT_BODY_PAGES=\number\numexpr\value{page}-1\relax}
\bibliographystyle{iclr2026_conference}
\bibliography{refs}

\clearpage
\appendix


\section{Proofs and computation of the certificates}\label{app:manifold}

This appendix proves the two lemmas that Section~\ref{subsec:followon} states:
Lemma~\ref{lem:isotropy} for the corrected curvature $\anisoH$, and
Lemma~\ref{lem:separable} for the residual $\crossx$. It then gives the
linear-algebra computation that turns the definition of $\crossx$ into a number,
extends the same manifold correction to the first-derivative test of
Section~\ref{subsec:hyp}, and calibrates these statistics against maps whose
class is known in advance.

\subsection{Proof of Lemma~\ref{lem:isotropy}}\label{app:proof-isotropy}

The lemma must show that a map which predicts exactly like a fixed-weight map on
the standardization manifold $\mathcal M$ can still differ from it in its second
derivatives, but only in a restricted way: the tangent curvature it can produce
is the same in every tangent direction. The idea is that such a map differs from
the fixed-weight map only through terms built out of the two functions whose
common zero set is $\mathcal M$, and each such term adds curvature only along the
two directions normal to $\mathcal M$, which the tangent projector $\PT$ removes,
except for a single scalar rescaling of $\PT$ itself that survives the removal.

Write $u(\ystdctx)=\mathbf 1^\top\ystdctx$ for the sum of the standardized labels
and $w(\ystdctx)=\lVert\ystdctx\rVert^2-r^2$ for the deviation of their squared
norm from $r^2$, so that $\mathcal M=\{u=0,\,w=0\}$ is exactly their common zero
set (Section~\ref{subsec:followon}). Since $\mathbf 1$ and $\ystdctx$ are linearly
independent whenever $r>0$, the gradients $\nabla u=\mathbf1$ and
$\nabla w=2\ystdctx$ are independent on $\mathcal M$, so $\mathcal M$ is a regular
level set of the pair $(u,w)$. If $g$ agrees on $\mathcal M$ with the affine map
$b+\tilde a^\top\ystdctx$, Hadamard's lemma applied in local coordinates adapted
to $(u,w)$ writes the difference between $g$ and this affine map as a combination
of $u$ and $w$ with smooth coefficient functions $p$ and $q$ defined near each
point of $\mathcal M$ \citep[Ch.~5 and Thm.~C.15]{lee2013smooth}:
\[
g=b+\tilde a^\top\ystdctx+p(\ystdctx)\,u(\ystdctx)+q(\ystdctx)\,w(\ystdctx).
\]
Because $u$ and $w$ vanish exactly on $\mathcal M$, this equation only restates
that $g$ and the affine map coincide there, while attaching a smooth multiplier
to each of the two ways of moving off $\mathcal M$.

At a point of $\mathcal M$, $u=w=0$ while their gradients need not vanish, so
differentiating gives
\begin{align*}
\nabla g&=\tilde a+p\,\mathbf 1+2q\,\ystdctx,\\
H&=2qI+\mathbf 1\nabla p^\top+\nabla p\,\mathbf 1^\top+2\ystdctx\nabla q^\top+2\nabla q\,\ystdctx^\top.
\end{align*}
The gradient of $g$ on $\mathcal M$ is therefore the fixed weight vector
$\tilde a$ plus a term along each of the two normal directions, and the Hessian
keeps the isotropic term $2qI$ and picks up four more terms, each proportional
to $\mathbf 1$ or to $\ystdctx$.

The normal space of $\mathcal M$ at $\ystdctx$ is spanned by $\mathbf 1$ and
$\ystdctx$, which are orthogonal because the standardized labels have zero mean,
$\mathbf1^\top\ystdctx=0$. The tangent projector
$\PT=I-\mathbf1\mathbf1^\top/n-\ystdctx\ystdctx^\top/r^2$ of
Section~\ref{subsec:followon} therefore has rank $n-2$ and annihilates both
normal directions, $\PT\mathbf1=\PT\ystdctx=0$. Projecting the Hessian above on
both sides removes all four extra terms, since each carries a factor of
$\mathbf 1$ or $\ystdctx$, and leaves
\[
\PT H\PT=2q\,\PT,
\]
which is the isotropic form the lemma claims, with $\lambda(\ystdctx)=2q(\ystdctx)$.
This proves Lemma~\ref{lem:isotropy}.

The intrinsic Riemannian Hessian of $g$ restricted to $\mathcal M$ is the
curvature computed from the values that $g$ takes on $\mathcal M$ alone, and not
from how $g$ extends off the manifold. The gradient identity above shows that it
differs from $\PT H\PT$ only by the isotropic correction
$-(\ystdctx^\top\nabla g/r^2)\PT$. Since this correction is itself proportional
to $\PT$, it is removed along with every other isotropic term once the trace is
subtracted, so the traceless part of $\PT H\PT$ equals the traceless part of the
intrinsic Hessian. Writing $\mathrm{traceless}(B)=B-\operatorname{tr}(B)\PT/(n-2)$
for a tangent-projected operator $B=\PT B\PT$, this traceless part is exactly
$T=\mathrm{traceless}(\PT H\PT)$ of Eq.~\eqref{eq:aniso}. Hence $T$ depends only
on the predictions that $g$ makes on $\mathcal M$ and not on how $g$ is defined
away from it, while the isotropic part $\lambda(\ystdctx)$ cannot be identified
from on-manifold behavior alone, since any value of $\lambda$ is compatible with
the same restriction of $g$ to $\mathcal M$.

\subsection{Proof of Lemma~\ref{lem:separable}}\label{app:proof-separable}

The lemma must show that the same argument extends to the larger class of maps
that transform each label separately before combining the results, so that the
corrected curvature such a map can produce always lies in the space
$\mathcal S=\{\mathrm{traceless}(\PT D\PT):D\text{ diagonal}\}$ of
Section~\ref{subsec:followon}, and the residual $\crossx$ of
Eq.~\eqref{eq:crossx} is zero.

Let $f=b+a^\top\ystdctx+\sum_k\phi_k(\ystd_k)$ agree with $g$ on $\mathcal M$,
where each $\phi_k$ acts on a single standardized label as in $g_{a,\phi}$
(Eq.~\eqref{eq:gaphi}). The difference $g-f$ vanishes identically on
$\mathcal M$, so applying the decomposition of the previous proof to $g-f$ shows
that its own tangent curvature is isotropic. The map $f$ itself produces the
tangent curvature $\PT D\PT$, where $D=\operatorname{diag}(\phi_k''(\ystd_k))$
collects the curvature of each $\phi_k$ and the linear term $a^\top\ystdctx$
contributes none. Adding this curvature back gives
\[
\PT H_g\PT=\PT D\PT+2q\,\PT
\]
for the tangent curvature of $g$ itself. Removing the trace cancels the isotropic
term $2q\PT$ and leaves $T=\mathrm{traceless}(\PT D\PT)$, which is a member of
$\mathcal S$ by definition, so $T\in\mathcal S$ and hence
$\crossx=\lVert T-P_{\mathcal S}T\rVert_F=0$.

As in the proof of Lemma~\ref{lem:isotropy}, the traceless projected Hessian of
any difference that vanishes on $\mathcal M$ is itself zero, so $T$, and with it
$\crossx$, does not depend on which off-manifold extension $g$ happens to have.

Both certificates are one-sided: a value of zero is consistent with the
corresponding class of maps, fixed-weight for $\anisoH$ and one-label-at-a-time
for $\crossx$, but it does not by itself prove that a model belongs to that
class, since some other map outside the class could in principle also give a zero
value on the labels tested.

\subsection{Computing the projection onto $\mathcal S$}\label{app:proof-projection}

Evaluating $\crossx$ requires the orthogonal projection $P_{\mathcal S}T$ of the
measured curvature $T$ onto $\mathcal S$, the space of corrected curvatures that
a one-label-at-a-time map can produce. Because $\mathcal S$ is spanned by one
operator per label, this projection reduces to a small linear solve rather than
a search over label maps.

For each label $k$, let $E_{kk}$ be the matrix with a single $1$ in position
$(k,k)$ and zeros elsewhere, so that a diagonal matrix $D$ is a combination of
the $E_{kk}$, and set $B_k=\mathrm{traceless}(\PT E_{kk}\PT)$ for the corrected
curvature that this single-label bump produces. The operators $B_1,\ldots,B_n$
therefore span $\mathcal S$.

Write $\beta=\operatorname{diag}(T)$ for the vector of diagonal entries of the
measured curvature. Because $T$ is already tangent-projected and traceless, its
inner product with each basis operator reduces to a single diagonal entry,
$\langle T,B_k\rangle=T_{kk}$, and the projection coefficients solve the normal
equations of this basis, $Gc=\beta$, where the Gram matrix of inner products
between basis operators is
\[
G_{jk}=(\PT)_{jk}^2-\frac{(\PT)_{jj}(\PT)_{kk}}{n-2}.
\]
The orthogonal projection is then $P_{\mathcal S}T=\sum_kc_kB_k$ with
$c=G^+\beta$, using the Moore--Penrose pseudoinverse because the basis is not
independent, and $\crossx=\bigl\lVert T-\sum_kc_kB_k\bigr\rVert_F$.

This solve stays well posed despite the rank deficiency of $G$. If $v$ lies in
the null space of $G$, the corresponding combination of basis operators vanishes,
$\sum_kv_kB_k=0$, so its inner product with $T$ is also zero, $v^\top\beta=0$,
and the system $Gc=\beta$ therefore remains consistent on the subspace where $G$
is singular. One such null direction is known in closed form, $G\mathbf1=0$,
because summing the $E_{kk}$ over every label gives the identity and
$\mathrm{traceless}(\PT I\PT)=\mathrm{traceless}(\PT)=0$. The pseudoinverse
handles this direction by returning the minimum-norm solution among those that
solve the system.

An independent implementation, which projects onto $\mathcal S$ using an
orthonormal basis rather than the $B_k$ above, gives the same residual to a
relative error of at most $2\times10^{-16}$ on the coupled control below and on
model Hessians of all five models. On the two null controls, both
implementations return values at the floating-point floor of about
$3\times10^{-14}$.

A nonzero $\crossx$ is possible only when $\mathcal S$ is a proper subspace of
the space in which $T$ lives. $\mathcal S$ is the image of the diagonal matrices
under $D\mapsto\mathrm{traceless}(\PT D\PT)$, so its dimension is at most $n-1$
because this map has a nontrivial null space. The traceless symmetric operators
on the tangent space, where $T$ lives, form a space of dimension
$(n-2)(n-1)/2-1$, which exceeds $n-1$ once $n\ge5$. For $n\ge5$, $\mathcal S$ is
therefore always a proper subspace, a condition met by a wide margin at the
context size $n=256$ that the paper uses for the certificates. For smaller
contexts, $\mathcal S$ fills the whole space at most label vectors but not at all
of them: at $n=4$ and $\ystdctx\propto(1,1,-1,-1)$, $\mathcal S$ has dimension
one inside a space of dimension two.

\subsection{The corrected perturbation test}\label{app:proof-perturbation}

Section~\ref{subsec:hyp} describes a first test for fixed-weight prediction:
perturb the context labels and compare the label sensitivities
$\ai=\partial g(\ystdctx)/\partial\ystd_i$ before and after.
Section~\ref{subsec:followon} shows that this test falls into the
standardization trap, as does the second-derivative test built from the
Hessian: a map can change its sensitivities under a permutation of the labels
while still making exactly the same predictions on every valid input. This section
defines the naive statistic precisely and repairs it with the same manifold
correction used for $\anisoH$ and $\crossx$.

Writing $a(\ystdctx)=\nabla_{\ystdctx}g(\ystdctx)$ for the vector collecting
these sensitivities and $P_\pi$ for the matrix that permutes the context labels
according to a permutation $\pi$, the naive permutation statistic averages the
relative change of this vector over queries $x_*$ and permutations $\pi$:
\begin{equation}
\permdelta=\mathbb E_{x_*,\pi}\!\left[\frac{\lVert a(P_\pi\ystdctx)-a(\ystdctx)\rVert_2}{\lVert a(\ystdctx)\rVert_2}\right].
\label{eq:permdelta}
\end{equation}
A fixed-weight map keeps $a$ unchanged under any permutation of the labels, so
$\permdelta=0$ for such a map exactly. But $\permdelta$ can be nonzero for a map
that is nonetheless a fixed-weight map on every valid input. For $g_a$
(Eq.~\eqref{eq:ga}), whose sensitivities are $a+2q\ystdctx$
(Section~\ref{subsec:followon}), a permutation changes them by
\[
a(P_\pi\ystdctx)-a(\ystdctx)=2q\,(P_\pi\ystdctx-\ystdctx),
\]
a vector that lies entirely in the span of $\ystdctx$ and $P_\pi\ystdctx$, and
$\permdelta$ reports this change even though $g_a$ makes the same predictions as
the fixed-weight map $b+a^\top\ystdctx$ everywhere it is ever evaluated.

The repair projects this span out before comparing sensitivities, the same
correction that $\PT$ performs for the second-derivative test. For each of six
permutations $\pi_k$, let $a_0=\nabla g(\ystdctx)$ and
$a_k=\nabla g(P_{\pi_k}\ystdctx)$ denote the sensitivities at the identity and
permuted labels, and let $Q_k$ project orthogonal to
$\operatorname{span}\{\mathbf1,\ystdctx,P_{\pi_k}\ystdctx\}$. The
intersection-projected displacement averages the residual change left after
this projection:
\[
\interdelta=\frac1{6}\sum_{k=1}^{6}\frac{\lVert Q_k(a_k-a_0)\rVert_2}{\lVert Q_ka_0\rVert_2+\eta}.
\]
A second statistic asks whether a single fixed weight vector can fit the
tangent-projected sensitivities at the identity and all six permutations at
once. With $P_k=\PT(P_{\pi_k}\ystdctx)$ the tangent projector evaluated at the
$k$-th permuted point, and $P_0$ the tangent projector at the identity, the
least-squares residual is
\[
\mathrm{LS}=\frac{\left(\min_{\hat a}\sum_{k=0}^{6}\lVert P_k(\hat a-a_k)\rVert_2^2\right)^{1/2}}{\left(\sum_{k=0}^{6}\lVert P_ka_k\rVert_2^2\right)^{1/2}+\eta}.
\]
Both statistics use the numerical guard $\eta=10^{-300}$ to keep the ratio
finite when its denominator is itself close to zero.

Both statistics vanish for every map that is affine on $\mathcal M$, by the
gradient identity from the proof of Lemma~\ref{lem:isotropy}: on $\mathcal M$,
the gradient of such a map differs from the fixed weight vector $\tilde a$ only
by components along $\mathbf1$ and $\ystdctx$, which at a permuted point become
components along $\mathbf1$ and $P_{\pi_k}\ystdctx$, exactly the directions that
$Q_k$ removes and that the single choice $\hat a=\tilde a$ absorbs once the
sensitivities are projected by $P_k$. Reported values average the per-query
statistics within each setting.

\subsection{Calibration on maps with known behavior}\label{app:proof-calibration}

We check the lemmas and the statistics above against three constructed maps
whose class is fixed by design, rather than only against the trained models, so
that a nonzero or a zero value can be read against a known answer before it is
trusted on a model whose class is unknown. All three controls pass through the
same estimator as the models, so the check exercises the same computation that
produces the numbers reported for the models and not a separate shortcut.

The affine control on $\mathcal M$ is a fixed-weight map, built like $g_a$
(Eq.~\eqref{eq:ga}). The separable control on $\mathcal M$ transforms one label
at a time, built like $g_{a,\phi}$ (Eq.~\eqref{eq:gaphi}). The coupled control
lets labels interact. Table~\ref{tab:immunity} reports all six statistics on the
three controls.

The two raw statistics are far from zero on the affine control: the naive
permutation statistic $\permdelta$ and the norm of the tangent curvature before
its isotropic part is removed. This is the standardization trap of
Section~\ref{subsec:followon} made numerical, since a fixed-weight map still
produces a large apparent change under these uncorrected measurements. The
corrected curvature $\anisoH$ removes this artifact and is zero, to within
floating-point error, on the affine control, but it is not zero on the separable
control, which does transform each label nonlinearly and so has genuine tangent
curvature. $\crossx$ is zero on both the affine and the separable control,
because Lemma~\ref{lem:separable} covers the affine control as the special case
$\phi_k=0$ of the affine-plus-separable class, and it is nonzero on the coupled
control, where every statistic in the table is nonzero.

The separable control's combination of a nonzero $\anisoH$ and a zero $\crossx$
is the pattern that the main claim of Section~\ref{sec:results-hessian} rests
on: a model can fail the fixed-weight test and still pass the one-label-at-a-time
test, and the calibration confirms that the estimator reports exactly this
pattern when it is the true answer. Float32 calibration puts a floor under the
zeros in the table, $\anisoH\le1.1\times10^{-5}$ on the affine null and
$\crossx\le9.3\times10^{-6}$ on the separable null, which corresponds to a share
of the squared curvature $\bar X_{\mathrm{sep}}$ (Eq.~\eqref{eq:shares}) of
$7\times10^{-11}$.

A separate, descriptive check perturbs pairs of labels directly rather than
computing the Hessian. Because a bump to one label that preserves the mean and
the norm of the labels changes the standardized values of other rows as well,
such pairwise perturbations can still respond to the separable control even
though it transforms only one label at a time, so they remain a descriptive
check and not a substitute for $\crossx$.

\begin{table}[t]\centering\small
\caption{\textbf{Calibration on maps with known behavior.} Values use the same
estimator as the models, at $n=256$ in float64. The raw statistics $\permdelta$
and $\lVert\PT H\PT\rVert_F$ can be nonzero even when the map is affine on
$\mathcal M$, which is the trap of Section~\ref{subsec:followon} made numerical.
Only $\crossx$ is also zero on a map that transforms one label at a time. The
first-order statistics LS and $\interdelta$ are zero on every map that is affine
on $\mathcal M$.}
\label{tab:immunity}
\begin{tabular}{lrrrrrr}
\toprule
Control & $\permdelta$ & $\lVert\PT H\PT\rVert_F$ & $\anisoH$ & $\crossx$ & LS & $\interdelta$ \\
\midrule
Affine on $\mathcal M$ & $1.21$ & $38.7$ & $4\!\times\!10^{-14}$ & $3\!\times\!10^{-14}$ & $6\!\times\!10^{-14}$ & $7\!\times\!10^{-14}$ \\
Separable on $\mathcal M$ & $1.23$ & $9.7$ & $1.11$ & $3\!\times\!10^{-14}$ & $0.36$ & $0.53$ \\
Coupled & $0.96$ & $0.73$ & $0.72$ & $0.72$ & $0.75$ & $0.96$ \\
\bottomrule
\end{tabular}
\end{table}

\section{Experimental details}\label{app:protocols}

This appendix documents the models, the datasets, and the Hessian computations
behind the results of Sections~\ref{sec:method} and~\ref{sec:results}, in the
order a reader needs them: which models we evaluate and where each one first
receives the context labels (Appendix~\ref{app:arms}), which datasets and how
many repetitions every experiment in the paper uses (Appendix~\ref{app:cells}),
and how we compute and validate the exact Hessians that the two certificates
depend on (Appendix~\ref{app:hvp}).

\subsection{Models and where the labels enter}\label{app:arms}

This subsection specifies the five models evaluated in
Section~\ref{sec:results-hessian}, the estimator inside each package whose output
our derivatives target, and the point in each architecture where the context
labels first enter the computation, so that the certificates of
Section~\ref{sec:method} can be traced back to a documented instrument.
Table~\ref{tab:arms} lists the five models together with the \emph{readout} we
differentiate in each one: the specific function of the network's output that
supplies the internal prediction $\yint$ of Section~\ref{subsec:hyp}. We differentiate this single readout inside each
package rather than its deployed default prediction, because that default
combines several such estimators, and Appendix~\ref{app:robustness} tests the
default predictions separately.

\begin{table}[t]\centering\small
\caption{\textbf{The five evaluated models and the readout we differentiate.}
The identifier names the checkpoint or model version that we load, and the
readout names the function of the network's output that supplies the internal
prediction $\yint$ rather than the package's default combined prediction.
Every derivative computation in this paper uses this single readout, except the readout-head check of Appendix~\ref{app:persetting}.}
\label{tab:arms}
\begin{tabular}{lll}
\toprule
Model & Identifier & Readout \\
\midrule
\TabICL{} \citep{tabiclv2} & \texttt{tabicl-regressor-v2} & linear mean-over-quantiles \\
\TabPFNiii{} \citep{tabpfn3} & \texttt{ModelVersion.V3} & bucket-softmax \\
\TabPFNvtwosix{} \citep{tabpfn25} & \texttt{ModelVersion.V2\_6} & bucket-softmax \\
\TabFM{} 1.0 \citep{tabfm} & \texttt{tabfm-1.0.0} (regression) & scalar regression output \\
\TabDPT{} v1.2 \citep{tabdptturbo} & \texttt{tabdpt1\_2.safetensors} & bucket-softmax \\
\bottomrule
\end{tabular}
\end{table}

The five models also differ in where the context labels first reach the
network, which matters for reading the results of
Section~\ref{sec:results-hessian}. \TabICL{} introduces the standardized label
vector at two points in its computation, once before the column-wise set
transformer that builds per-feature representations from the context, and once
more at the in-context stage that forms the prediction from those
representations, so that the same label vector we differentiate feeds both sites
in a full forward pass. \TabPFNiii{} follows the same two-stage design: it adds
an encoding of each context label to the cell embeddings of that row before the
attention that builds the row representations, and a second encoding before its
in-context stage, and the same label vector feeds both sites. \TabPFNvtwosix{}
injects the labels once, as an additional column that its alternating attention
across features and rows then shares across the context. According to its released code \citep{tabfmcode}, \TabFM{} also
introduces the labels at two points: it adds an encoding of each context label to
every feature cell of that row before attention across features and rows, and a
second encoding to the compressed row representation before the in-context
stage, and the same label vector feeds both sites. Released
\TabDPT{} v1.2 \citep{tabdptturbo} instead re-encodes the labels into the values that attention
combines at every layer, so its queries and keys receive no direct label
injection even though its deeper residual streams still carry label-derived
information indirectly. This description holds for the released \TabDPT{} v1.2
inference code that we evaluate.

We run each package with its default feature preprocessing for a single
estimator, except that we turn off the feature normalization of \TabFM{}, and we
run every derivative computation in \texttt{float32}. We differentiate with respect to the internal
standardized label vector $\ystdctx$ of Section~\ref{subsec:hyp}, and we treat
the output conversion back to raw label units as a fixed map that plays no part
in the derivative. Package versions, constructors, runtime settings, and the
digests that identify the exact model state loaded for each experiment are
recorded in the supplementary material.

Forward passes and first-order derivatives use each package's own attention
computation, with mixed precision and the optional fused kernels of \TabICL{} and
\TabDPT{} turned off, and the Hessian experiments select the standard, non-fused
attention backend for the whole run. Second derivatives need an attention
implementation with a second backward pass, so every Hessian in this paper
substitutes standard (eager) softmax attention, and Appendix~\ref{app:hvp}
validates this substitution against the deployed kernels directly. Under eager
attention, the first derivatives of the readout match those computed with the
native kernels to a relative error of at most $8\times10^{-5}$ for all five
models, and the forward prediction of \TabFM{} matches its native prediction to
$1.2\times10^{-5}$ standard deviations of the output. 

\subsection{Datasets and protocol}\label{app:cells}

This subsection lists the datasets behind every experiment in the paper and the
grid and replication each one uses, so that Table~\ref{tab:cells} serves as a
single reference for what a given result covers and how many times we repeat
the measurement.

\paragraph{Datasets.} The main experiment of Section~\ref{sec:results-hessian}
evaluates four datasets: a synthetic regression task we call
\texttt{synthetic\_mi} (the generator paragraph below), and the OpenML datasets
\texttt{kin8nm} (189), \texttt{pol} (201), and \texttt{superconduct} (43174). An
eight-dataset \emph{core} grid extends this set with \texttt{ailerons} (296),
\texttt{abalone} (44956), \texttt{concrete} (44959), and \texttt{house\_16H}
(574), and supplies the settings for the first-order tests and for
the surrogate-replacement panel, except that the seven-dataset surrogate panel
excludes \texttt{abalone}.

Table~\ref{tab:cells} records the grid and the replication of every experiment
in the paper, writing \emph{Stat 4} for the four-dataset main grid
above, \emph{Core 8} for the eight-dataset grid, and \emph{Core 7} for its
seven-dataset restriction, each crossed with the stated number of models. The
rows with four models do not include \TabFM{}, which enters only the
exact-Hessian experiments and the Hessian validation of Appendix~\ref{app:hvp}.

\begin{table}[t]\centering\small
\caption{\textbf{Grid and replication of every experiment in the paper.}
Each row names an experiment and gives its grid (the datasets and models it
covers) and its replication. A \emph{setting} is one model evaluated
on one dataset. \emph{Stat 4}, \emph{Core 8}, and \emph{Core 7} are the dataset
grids defined in the text. $n$ is the number of context rows, $n_q$ the number
of query points, $M$ the number of label draws used to fit a surrogate, and
\emph{probes} the number of random tangent directions used to estimate a
Hessian statistic without forming the full matrix (Appendix~\ref{app:hvp}).}
\label{tab:cells}
\begin{tabular}{l>{\raggedright\arraybackslash}p{0.26\linewidth}>{\raggedright\arraybackslash}p{0.38\linewidth}}
\toprule
Experiment & Grid and settings & Replication \\
\midrule
First-order tests & Core 8, 4 models, 32 & $n=256$, 3 contexts, $n_q=8$, 6 permutations \\
Projected Hessian & Stat 4, 4 models, 16 & $n_q=4$, 3 contexts, 16 paired probes \\
Probe-based certificates & Stat 4, 4 models, 16 & $n=256$, 3 contexts, 64 probes, $n_q=8$ \\
Forward second differences & Stat 4, 4 models, 16 & $n=256$, 3 contexts, first query, 4 tangent directions per context, 3 step sizes \\
Exact Hessian & Stat 4, 5 models, 20 & $n=256$, 3 contexts, $n_q=4$ \\
Hessian validation & 2 datasets, 5 models, 10 & $n=256$ \\
Readout head & 2 datasets, 2 models, 4 & $n=256$, 2 contexts, $n_q=2$ \\
Default predictions & 2 datasets, 3 models, 6 & 2 contexts, and the feature orderings of \TabDPT{} on the same 2 datasets \\
Surrogate replacement & Core 7, 3 models, 21 & $n=96$, $M=400$, 300 test queries \\
Observed-feature comparator & Core 7, 4 models, 28 & $n=96$, 300 test queries \\
Task-preserving surrogate & 3 datasets, 4 models, 12 & $n=64$, $n_q=6$, $M=210$ (wild bootstrap: \texttt{kin8nm}, \texttt{pol}, \texttt{superconduct}) \\
Constructed local model & 3 datasets, 3 models, 9 & $n=128$, $n_q=5$, 24 label draws (construction: 3 synthetic tasks at $n=96$) \\
Context size, $R_T$ & 2 datasets, 3 models, 6 & $n=128$--$1024$, 1 context, $n_q=2$, 8 probes \\
Context size, $\bar X_{\mathrm{sep}}$ & 2 datasets, 2 models, 4 & $n=128$, $256$, $512$, 2 contexts, $n_q=2$ \\
Random copies & 2 datasets, 5 models, 10 & 12 copies per model, each evaluated on both datasets \\
Random copies, $\permdelta$ & 2 datasets, 4 models, 8 & $n=128$ (\TabICL{}: $256$), 12 copies per model \\
Attention-entropy control & 2 datasets, 2 models, 4 & 3 copies per model, each evaluated on both datasets \\
All-layer route patch & Stat 4, 4 models, 16 & $n=256$, $n_q=8$, 6 permutations \\
Controlled models & Gaussian-process task & 3 seeds, and a five-seed extension of 75 runs \\
\bottomrule
\end{tabular}
\end{table}

\paragraph{Pretraining overlap.} The evaluated \TabDPT{} v1.2 checkpoint
(TabDPT-Turbo) was pretrained on $1{,}445$ OpenML datasets, from which its authors
removed possible duplicates of the CC18, CTR23, and TabArena benchmarks, and no
list of these datasets is released \citep{tabdptturbo}. We therefore cannot
check which of our real datasets it saw during pretraining. The $123$-dataset
manifest of the original \TabDPT{} \citep{tabdpt} contains \texttt{ailerons} directly and
\texttt{pol} and \texttt{house\_16H} only through classification variants. Such
an overlap does not change the measured response of the evaluated map, but it
limits how far \TabDPT{}'s results on these datasets can be read as evidence of
generalization independent of pretraining.

\paragraph{Metrics.}\label{app:metrics}
The sensitivities $\effw$ that the certificates use are local derivatives, and
they describe a rate of change rather than an additive contribution to a finite
prediction change. The first-order tests permute the standardized labels inside
the model, so the context rows stay aligned without any further check. The surrogate experiments of Appendix~\ref{app:tuning} fit an affine surrogate
to a model's predictions and evaluate it on label draws held out from fitting.
The surrogate-replacement panel fits it on $400$ label permutations, choosing
the ridge regularization from a grid spanning $10^{-3}$ to $10^{2}$ on an 85/15
split of these draws, and evaluates it on $100$ fresh permutations. The
task-preserving surrogate fits it on $210$ of $300$ label draws, choosing the
regularization from a grid spanning $10^{-3}$ to $10^{3}$ by leave-one-out
cross-validation, and evaluates it on the remaining $90$.
The task-preserving draws of that appendix change the moments of the labels and
are fitted in raw label coordinates, unlike the derivative measurements
elsewhere in this appendix, which use standardized labels. Because the fit uses
finitely many draws and a regularized estimator, a surrogate $R^2$ below $1$
does not by itself establish that the residual gap reflects the underlying map
rather than fitting noise or the chosen regularization.

\paragraph{Synthetic generator.}\label{app:generator}
Features follow $x\sim\mathcal U[0,1]^{16}$. Rows of
$A\in\mathbb R^{n_{\mathrm{idx}}\times16}$ are drawn from a standard
normal. The experiments on the certificate grid of Table~\ref{tab:crossx} scale
each row to unit Euclidean norm. These are the exact and probe-based
certificates, the first-order tests, the forward second differences, the
readout-head check, the context-size check of $\bar X_{\mathrm{sep}}$, the
default-prediction check, the certificates of the random copies, and the Hessian
validation on \TabICL{} and \TabFM{}. The other experiments, including the
context-size check of $R_T$, divide $A$ by its largest column sum of absolute
values. The target is
\[
f(x)=\sum_{k=1}^{n_{\mathrm{idx}}}\sum_{j=1}^{6}
w_{kj}\exp\!\left[-\frac{(A_k^\top x-c_{kj})^2}{2(0.2)^2}\right],
\qquad y=f(x)+0.05\varepsilon,
\]
where $w_{kj},\varepsilon\sim\mathcal N(0,1)$ and
$c_{kj}=0.5\sum_iA_{ki}+u_{kj}$ with $u_{kj}\sim\mathcal U(-0.6,0.6)$.
The core task uses four indices.

\subsection{Computing and validating the Hessians}\label{app:hvp}

Second derivatives need attention implemented with eager softmax, because the
fused attention kernels that packages deploy do not support a second backward
pass (Appendix~\ref{app:arms}), so every exact Hessian in this paper comes from
a different computational path than the predictions and first-order
sensitivities that accompany it. Before treating these Hessians as evidence, we
must show that substituting eager attention at this one step leaves the computed
derivatives unchanged, which is what this subsection validates.

\paragraph{Assembling and estimating the Hessians.} We assemble each exact
Hessian one column at a time. Differentiating the full sensitivity vector
$a=\nabla_{\ystdctx}g(\ystdctx)$ with respect to a single label $\ystd_j$ gives
$He_j=\partial a/\partial\ystd_j$, the $j$-th column of $H_{\ystdctx}$, where
$e_j$ is the $j$-th standard basis vector in $\mathbb R^n$, and stacking its $n$
columns recovers the exact Hessian. We compute each column twice on the same
contexts, once from the full network and once with derivatives blocked from
passing through the attention scores, the intervention of
Section~\ref{sec:results-origin}.

Forming every column of a full Hessian at every setting in the paper would be
costly to repeat, so several of the appendix analyses that do not need the exact
matrix instead estimate $\anisoH$ from a small set of shared tangent Rademacher
probes. These probes are random tangent-space directions with independent
$\pm1$ entries, and we use $8$, $16$ or $64$ of them per setting to estimate
the projected norm and the trace of the tangent curvature $\PT H_{\ystdctx}\PT$. The squared
trace uses a split-product estimator, which multiplies two independent
probe-based estimates of the trace rather than squaring one, so that the noise
of a single estimate does not enter as a bias. Because a probe-based estimate of
$\anisoH$ can occasionally fall outside the range that a true squared norm
allows, we clamp it to its admissible range, which introduces a small null bias
in return: in a $64$-probe simulation of the affine control $g_a$
(Eq.~\ref{eq:ga}), this bias has mean $0.48$ and $95$th percentile $1.60$, small
against the projected norm of $38.7$ measured in the same simulation. The exact
Hessians reported in Table~\ref{tab:main} do not use this clamp. We also
re-center the standardized labels inside each tangent projector, which
compensates for small floating-point differences between how different packages
implement the standardization wrapper of Section~\ref{subsec:hyp}.

\paragraph{Validating the eager-attention path.} The functions we differentiate
compose linear maps, softmax, smooth activations (GELU, SiLU), and normalization
layers (LayerNorm, RMSNorm) away from their degenerate points, and the discrete steps of the standardization wrapper sit
outside the standardized-label leaf that these derivatives are taken against,
so nothing discontinuous enters the derivative itself. We compare the exact
Hessian-vector products computed under eager attention with central finite
differences of the gradient returned by each package's native fused kernel:
\[
\frac{a(\ystdctx+\varepsilon v)-a(\ystdctx-\varepsilon v)}{2\varepsilon},
\]
where $v$ is a probe direction and $\varepsilon$ the finite-difference step
size. We use the same seeded Rademacher directions for both computations,
together with unit-vector directions that validate individual Hessian columns
directly, and we select the step $\varepsilon^\star$ at a plateau in the
relative error where adjacent step sizes give converged answers.

At $n=256$, every one of the ten model--dataset settings we tested meets our
validation thresholds: cosine similarity of at least $0.99$, relative error of
at most $0.05$, and a matched-column fraction of at least $0.90$ between the two
computations. The agreement is in practice closer than these thresholds
require. Every median cosine rounds to $1.0000$, the highest median relative
error is $0.0034$, every matched-column fraction reaches $1.00$, and the selected steps
are $10^{-3}$ or $3\times10^{-3}$, except for \TabFM{}, whose \texttt{float32}
finite differences become noisy below $10^{-2}$. Table~\ref{tab:hvp} reports these ten
settings, and the supplementary material holds the full validation and probe
records.

\begin{table}[t]\centering\small
\caption{\textbf{Validating the eager-attention Hessians against the native
fused-kernel gradient}, at $n=256$ on \texttt{synthetic\_mi} and
\texttt{kin8nm}, two of the four datasets in the main grid. For each
model--dataset setting, $\varepsilon^\star$ is the selected finite-difference
step, cos and rel-$L^2$ are the cosine similarity and relative $L^2$ error
between the eager Hessian-vector product and the central finite difference of
the native gradient along the same probe direction, reported as the median over
probes and queries with the maximum in parentheses, matched is the fraction of
Hessian columns on which the two computations agree, and
$\lVert H\rVert_F/\lVert a\rVert$ gives the scale of the Hessian relative to the
sensitivity vector for context, not a validation criterion. Every setting meets
the thresholds stated in the text.}
\label{tab:hvp}
\begin{tabular}{llccccc}
\toprule
Model & Dataset & $\varepsilon^\star$ & cos & rel-$L^2$ med (max) & matched & $\lVert H\rVert_F/\lVert a\rVert$ \\
\midrule
\TabICL{}   & \texttt{synthetic\_mi} & $3\!\times\!10^{-3}$ & $1.0000$ & $0.0008\ (0.0012)$ & $1.00$ & $5.51$ \\
\TabICL{}   & \texttt{kin8nm}        & $3\!\times\!10^{-3}$ & $1.0000$ & $0.0010\ (0.0020)$ & $1.00$ & $4.62$ \\
\TabPFNiii{}& \texttt{synthetic\_mi} & $3\!\times\!10^{-3}$ & $1.0000$ & $0.0014\ (0.0046)$ & $1.00$ & $7.29$ \\
\TabPFNiii{}& \texttt{kin8nm}        & $3\!\times\!10^{-3}$ & $1.0000$ & $0.0020\ (0.0063)$ & $1.00$ & $4.99$ \\
\TabPFNvtwosix{}     & \texttt{synthetic\_mi} & $3\!\times\!10^{-3}$ & $1.0000$ & $0.0004\ (0.0006)$ & $1.00$ & $3.02$ \\
\TabPFNvtwosix{}     & \texttt{kin8nm}        & $3\!\times\!10^{-3}$ & $1.0000$ & $0.0004\ (0.0005)$ & $1.00$ & $3.14$ \\
\TabFM{}     & \texttt{synthetic\_mi} & $1\!\times\!10^{-2}$ & $1.0000$ & $0.0024\ (0.0033)$ & $1.00$ & $22.1$ \\
\TabFM{}     & \texttt{kin8nm}        & $1\!\times\!10^{-2}$ & $1.0000$ & $0.0034\ (0.0065)$ & $1.00$ & $5.78$ \\
\TabDPT{}    & \texttt{synthetic\_mi} & $1\!\times\!10^{-3}$ & $1.0000$ & $0.0003\ (0.0004)$ & $1.00$ & $8.40$ \\
\TabDPT{}    & \texttt{kin8nm}        & $1\!\times\!10^{-3}$ & $1.0000$ & $0.0004\ (0.0006)$ & $1.00$ & $6.93$ \\
\bottomrule
\end{tabular}
\end{table}

\section{Additional results}\label{app:additional}

\subsection{Certificates in every setting}\label{app:persetting}

This subsection supports Table~\ref{tab:main} of the main text
(Section~\ref{sec:results-hessian}), which reports each certificate averaged
over four datasets per model. It adds three things to that table: (1) the
per-setting values behind its means, (2) a first-order version of the
fixed-weight test that does not depend on differentiating the sensitivities
twice, and (3) checks that the interaction found by the second-order
certificates is not an artifact of the readout head or of the exact-Hessian
instrument used to compute it. It closes with a look at how the certificate
behaves at context sizes other than the $\nctx=256$ rows used in the main
experiments.

\paragraph{First-order tests.} A first-order counterpart to the fixed-weight
test avoids differentiating the sensitivities twice.
Appendix~\ref{app:manifold} defines the intersection-projected displacement
$\interdelta$ and the least-squares residual LS, two statistics built from the
tangent gradient alone that are zero for every map that agrees with a
fixed-weight map on $\mathcal M$. Both reject the affine null on all $32$ core
settings, with LS residuals between $0.722$ and $0.925$, which shows that the
departure from fixed weights does not depend on the second-order instrument.

\begin{table}[t]\centering\small
\caption{\textbf{First-order and probe-based certificate results, by setting.}
Each row averages three sampled contexts
and eight queries per setting at $\nctx=256$. The columns
$\anisoH/\lVert\PT a\rVert$, $\bar A_T$, and Att. drop are estimated from tangent
probes (Appendix~\ref{app:hvp}). $\anisoH/\lVert\PT a\rVert$ scales the
certificate by the norm of the tangent gradient, so that models with larger raw
sensitivities do not appear to have a proportionally larger certificate. Att. drop
is the relative fall of $\anisoH$ when derivatives cannot pass through the
attention scores (Section~\ref{sec:results-origin}), and the brackets enclose
the three per-context paired-probe $95\%$ bootstrap intervals. $\interdelta$
and LS are the corrected first-order statistics of
Appendix~\ref{app:manifold}, which likewise vanish for every map that is affine
on $\mathcal M$.}
\label{tab:manifold}
\begin{tabular}{llrrrrr}
\toprule
Model & Dataset & $\anisoH/\lVert\PT a\rVert$ & $\bar A_T$ & Att. drop [interval] & $\interdelta$ & LS \\
\midrule
\TabICL{} & \texttt{synthetic\_mi} & 5.08 & 0.991 & 0.874 {\scriptsize[$0.82,0.93$]} & 1.08 & 0.893 \\
\TabICL{} & \texttt{kin8nm} & 4.40 & 0.993 & 0.814 {\scriptsize[$0.74,0.86$]} & 1.02 & 0.865 \\
\TabICL{} & \texttt{pol} & 35.75 & 0.972 & 0.874 {\scriptsize[$0.79,0.92$]} & 1.12 & 0.783 \\
\TabICL{} & \texttt{superconduct} & 10.05 & 0.987 & 0.749 {\scriptsize[$0.68,0.82$]} & 0.99 & 0.789 \\
\TabPFNiii{} & \texttt{synthetic\_mi} & 8.76 & 0.996 & 0.933 {\scriptsize[$0.88,0.97$]} & 1.01 & 0.915 \\
\TabPFNiii{} & \texttt{kin8nm} & 4.82 & 0.993 & 0.850 {\scriptsize[$0.81,0.91$]} & 1.00 & 0.907 \\
\TabPFNiii{} & \texttt{pol} & 57.03 & 0.960 & 0.882 {\scriptsize[$0.84,0.91$]} & 1.52 & 0.885 \\
\TabPFNiii{} & \texttt{superconduct} & 36.32 & 0.996 & 0.880 {\scriptsize[$0.72,0.96$]} & 1.14 & 0.907 \\
\TabPFNvtwosix{} & \texttt{synthetic\_mi} & 3.14 & 0.988 & 0.912 {\scriptsize[$0.86,0.95$]} & 1.21 & 0.899 \\
\TabPFNvtwosix{} & \texttt{kin8nm} & 2.71 & 0.985 & 0.849 {\scriptsize[$0.83,0.87$]} & 1.10 & 0.885 \\
\TabPFNvtwosix{} & \texttt{pol} & 16.30 & 0.949 & 0.964 {\scriptsize[$0.95,0.97$]} & 2.52 & 0.922 \\
\TabPFNvtwosix{} & \texttt{superconduct} & 7.51 & 0.989 & 0.932 {\scriptsize[$0.91,0.95$]} & 1.38 & 0.901 \\
\TabDPT{} & \texttt{synthetic\_mi} & 9.27 & 0.994 & 0.840 {\scriptsize[$0.75,0.89$]} & 1.00 & 0.925 \\
\TabDPT{} & \texttt{kin8nm} & 6.71 & 0.995 & 0.762 {\scriptsize[$0.73,0.81$]} & 1.01 & 0.922 \\
\TabDPT{} & \texttt{pol} & 31.04 & 0.948 & 0.655 {\scriptsize[$0.50,0.74$]} & 1.33 & 0.810 \\
\TabDPT{} & \texttt{superconduct} & 10.32 & 0.991 & 0.568 {\scriptsize[$0.50,0.61$]} & 0.99 & 0.879 \\
\bottomrule
\end{tabular}
\end{table}

\paragraph{Shares of the squared curvature in every setting.} The exact
normalized certificate $\bar X_{\mathrm{sep}}$ is $0.239$ to $0.916$ across the
$20$ settings, with a mean of $0.63$ (Table~\ref{tab:crossx}). The exact
$\bar A_T$ ranges from $0.952$ to $0.997$. On the $16$ settings where both
instruments were run, it ranges from $0.952$ to $0.996$ and closely matches the
cheaper probe estimate of the same quantity, $0.948$ to $0.996$, which checks
that the probe-based instrument used in Table~\ref{tab:manifold} agrees with the
exact Hessian used elsewhere.

\begin{table}[t]\centering\small
\caption{\textbf{Exact certificate of joint processing, by setting.} Behind the
per-model means of Table~\ref{tab:main}, each row reports $\bar X_{\mathrm{sep}}$ for one setting as the
mean over three sampled contexts and four queries at $\nctx=256$, with the
range of the three context means (Range). This certificate is exactly zero for
every map that is affine plus one-label-at-a-time on $\mathcal M$
(Section~\ref{sec:method}), and its float32 implementation floors it at
$7\times10^{-11}$. $\bar A_T$ also comes from the exact Hessian, whereas Table~\ref{tab:manifold} estimates it from probes. $\crossx/\lVert\PT a\rVert$ scales
$\crossx$ by the norm of the tangent gradient. The two Att. drop columns give
$1-\anisoH(H_{\mathrm{det}})/\anisoH(H)$ and $1-\crossx(H_{\mathrm{det}})/\crossx(H)$, the
relative falls of $\anisoH$ and $\crossx$ when derivatives cannot pass through the
attention scores.}
\label{tab:crossx}
\setlength{\tabcolsep}{4.5pt}
\begin{tabular}{ll r c r r r r}
\toprule
 & & & & & & \multicolumn{2}{c}{Att. drop} \\
\cmidrule(lr){7-8}
Model & Dataset & $\bar X_{\mathrm{sep}}$ & Range & $\bar A_T$ & $\crossx/\lVert\PT a\rVert$ & $\anisoH$ & $\crossx$ \\
\midrule
\TabICL{} & \texttt{synthetic\_mi} & $0.864$ & $[0.836,0.900]$ & $0.990$ & $4.80$ & $0.878$ & $0.895$ \\
\TabICL{} & \texttt{kin8nm} & $0.772$ & $[0.700,0.813]$ & $0.993$ & $4.17$ & $0.820$ & $0.848$ \\
\TabICL{} & \texttt{pol} & $0.322$ & $[0.306,0.353]$ & $0.977$ & $20.7$ & $0.885$ & $0.838$ \\
\TabICL{} & \texttt{superconduct} & $0.566$ & $[0.552,0.575]$ & $0.990$ & $5.94$ & $0.681$ & $0.700$ \\
\TabPFNiii{} & \texttt{synthetic\_mi} & $0.888$ & $[0.844,0.922]$ & $0.996$ & $8.49$ & $0.935$ & $0.968$ \\
\TabPFNiii{} & \texttt{kin8nm} & $0.745$ & $[0.721,0.770]$ & $0.994$ & $5.25$ & $0.864$ & $0.919$ \\
\TabPFNiii{} & \texttt{pol} & $0.496$ & $[0.439,0.530]$ & $0.953$ & $36.1$ & $0.783$ & $0.830$ \\
\TabPFNiii{} & \texttt{superconduct} & $0.702$ & $[0.662,0.757]$ & $0.995$ & $30.5$ & $0.852$ & $0.884$ \\
\TabPFNvtwosix{} & \texttt{synthetic\_mi} & $0.837$ & $[0.772,0.896]$ & $0.981$ & $2.82$ & $0.891$ & $0.952$ \\
\TabPFNvtwosix{} & \texttt{kin8nm} & $0.774$ & $[0.741,0.818]$ & $0.982$ & $2.36$ & $0.857$ & $0.931$ \\
\TabPFNvtwosix{} & \texttt{pol} & $0.567$ & $[0.547,0.599]$ & $0.954$ & $12.7$ & $0.962$ & $0.968$ \\
\TabPFNvtwosix{} & \texttt{superconduct} & $0.758$ & $[0.739,0.788]$ & $0.990$ & $6.48$ & $0.926$ & $0.961$ \\
\TabFM{} & \texttt{synthetic\_mi} & $0.916$ & $[0.882,0.977]$ & $0.995$ & $11.4$ & $0.921$ & $0.941$ \\
\TabFM{} & \texttt{kin8nm} & $0.829$ & $[0.817,0.843]$ & $0.991$ & $5.25$ & $0.893$ & $0.928$ \\
\TabFM{} & \texttt{pol} & $0.251$ & $[0.242,0.260]$ & $0.985$ & $30.4$ & $0.581$ & $0.628$ \\
\TabFM{} & \texttt{superconduct} & $0.703$ & $[0.672,0.718]$ & $0.997$ & $8.43$ & $0.734$ & $0.789$ \\
\TabDPT{} & \texttt{synthetic\_mi} & $0.447$ & $[0.366,0.553]$ & $0.992$ & $6.37$ & $0.840$ & $0.914$ \\
\TabDPT{} & \texttt{kin8nm} & $0.435$ & $[0.417,0.446]$ & $0.996$ & $4.32$ & $0.785$ & $0.899$ \\
\TabDPT{} & \texttt{pol} & $0.239$ & $[0.174,0.292]$ & $0.952$ & $16.3$ & $0.550$ & $0.527$ \\
\TabDPT{} & \texttt{superconduct} & $0.496$ & $[0.423,0.567]$ & $0.992$ & $7.37$ & $0.546$ & $0.673$ \\
\bottomrule
\end{tabular}
\end{table}

\paragraph{The interaction is present before the readout head.} Three of the
models, the two \TabPFN{} models and \TabDPT{}, turn the network's logits over a
grid of label bins into a prediction through a softmax (the bucket-softmax
readout of Table~\ref{tab:arms}). Such a head is itself a nonlinear map of the
logits, so it can contribute curvature even when the logits are an affine
function of the labels, through the chain-rule term
$J_\ell^\top(\nabla^2 h)J_\ell$, where $J_\ell$ is the Jacobian of the logits
with respect to the labels and $h$ is the head. Because this term would inflate
the certificates for a reason unrelated to joint label processing inside the
network, we recompute both certificates from a frozen linear functional of the
pre-head logits, which removes the term. On the two \TabPFN{} models, evaluated
on two datasets and two contexts each, all eight context-settings keep nonzero
certificates under this readout, with $\bar X_{\mathrm{sep}}$ between $0.50$ and
$0.97$. Their ratio to the values obtained with the mean readout is
$0.95\pm0.13$, with a minimum of $0.70$, which places most of the measured
interaction upstream of the bucket head on these settings.

\paragraph{The differentiated curvature matches the curvature of actual
predictions.} The certificates above come from differentiating the internal
prediction at the observed labels. As a check that this differentiated
curvature matches the curvature that the network actually shows along the
manifold, we compare it with a purely forward quantity: the second difference
of the prediction itself, evaluated at points of $\mathcal M$ near the observed
labels. For a tangent direction $v$, the second derivative of a prediction
function $F$ along the renormalized label path is
\begin{equation}
B_F(v)=v^\top H_Fv-\frac{\lVert v\rVert^2}{r^2}\,\ystdctx_c^\top\nabla F,
\label{eq:intrinsic-directional-appendix}
\end{equation}
where $\ystdctx_c$ is the centered standardized-label vector, $H_F$ and
$\nabla F$ are the Hessian and the gradient of $F$, and the second term removes
the curvature that the path picks up from staying on the sphere, the same
correction that builds $T$ in Section~\ref{sec:method}. We compare $B_F(v)$ with
central differences of the prediction along four unit tangent directions per
context at the first query, for the four models other than \TabFM{}, with steps of $0.05$, $0.1$, and $0.2$. The full
internal readout reaches a median setting-level relative error of $0.046$,
and $15$ of the $16$ settings fall below the descriptive threshold of $0.5$,
with only \TabDPT{} on \texttt{pol} above it at $0.567$.

\paragraph{Context-size dependence.} Because the certificates in
Table~\ref{tab:main} use $\nctx=256$ context rows, we also check whether the
interaction strengthens or weakens with more context. For this check we use
the ratio $R_T=\lVert\PT H\PT\rVert_F/\lVert\PT a\rVert$. Unlike $\anisoH$,
this ratio keeps the isotropic part of the tangent curvature and simply scales
its norm by the tangent gradient norm. For \TabPFNiii{}, \TabPFNvtwosix{} and
\TabDPT{} on \texttt{synthetic\_mi} and \texttt{kin8nm}, $R_T$ varies by at most a factor of $2.4$ and shows no consistent
trend from $128$ to $1024$ rows, so the same ratio divided by $\sqrt{n}$
declines, and this descriptor does not show a strengthening of the curvature
with context size. Two of the largest-context \TabPFNvtwosix{} settings exceeded
available memory.
For \TabICL{} and \TabPFNvtwosix{} on the same two datasets, the exact
normalized certificate $\bar X_{\mathrm{sep}}$ ranges over the context means,
with two contexts and two queries at each size, from $0.80$ to $0.96$ at
$n=128$, $0.72$ to $0.91$ at $n=256$, and $0.67$ to $0.91$ at $n=512$, a mild
decline that leaves the certificate present at every tested size.

\subsection{The default predictions are not affine in the labels}\label{app:robustness}

The two certificates of Section~\ref{sec:method} are computed on a single
internal estimator inside each package, since this is the quantity for which
the network's own derivatives are well defined (Section~\ref{subsec:arms}). By
contrast, each package returns a complete prediction that averages several such
estimators together. This subsection checks separately whether that complete
prediction is also non-affine in the labels, the check that
Section~\ref{sec:results-hessian} of the main text cites.

\paragraph{Construction.} The test perturbs the labels along two orthogonal
tangent bumps $\delta_i$ and $\delta_j$, each of scale $\varepsilon$, chosen so
that $\delta_i^\top\delta_j=0$. Starting from the standardized labels
$\ystdctx$, a step $v$ is applied through the exactly renormalized path
$m(v)=\overline{\ystd}\mathbf 1+(\ystdctx_c+v)r/\lVert\ystdctx_c+v\rVert$,
which keeps the mean and the norm of the labels fixed at every step, so that
$m(v)$ always lies on the standardization manifold $\mathcal M$ of
Section~\ref{subsec:followon}. The test evaluates the four-point difference
\[
D_{ij}=g(m(\delta_i+\delta_j))-g(m(\delta_i))-g(m(\delta_j))+g(\ystdctx).
\]
For a map that is affine on $\mathcal M$, the order-$\varepsilon^2$ term that
the four-point difference could otherwise pick up from the curved path is
proportional to $\delta_i^\top\delta_j$, so the chosen orthogonality sets this
term to zero and the affine null becomes $O(\varepsilon^3)$. The orthogonality
of $\delta_i$ and $\delta_j$ is required for this cancellation, and a
one-label-at-a-time map can still produce a nonzero response under this test.
The native test applies the affine image of these paths in raw-label
coordinates and evaluates the complete prediction returned by each package's
default inference procedure, with both the label mean and the label scale held
fixed.

\paragraph{Results.} We calibrate the test on setting-specific controls: an
affine map, and an impostor that agrees with an affine map on the fixed-moment
set but not outside it, in the manner of $g_a$. A third control, a map that lets
the labels interact, checks that the test fires. With the member seeds held
fixed, the measured interaction exceeds the floor set by the first two controls
by a factor of $44$ to $16{,}900$ on $12$ context-settings, spanning three
models, two datasets, and two contexts. We had required in advance that this
floor stay below $5\%$ of the interaction of the third control. The two
\texttt{kin8nm} contexts miss this requirement, with floors at $6$ to $9\%$, and
they give the smallest factors, $44$ to $279$, which still place the measured
interactions far above the floor. Averaged over the two contexts of each model and
dataset, the ratio between the complete and the single-estimator response is
$0.41$ to $1.03$ ($0.30$ to $1.37$ for single contexts). \TabDPT{}
additionally averages over several feature orderings, and this separate
averaging axis exceeds its calibrated floor by a factor of $55$ to $616$ on four
context-settings, with ratios of $0.54$ to $1.54$ between the averaged and the
single response.

\paragraph{Scope.} These functional checks support the finding that the default
predictions are not affine in the labels. They do not exclude a complete
prediction that transforms each label one at a time and combines the results
without interaction, since such a map is also non-affine. The exact certificate
$\crossx$ therefore remains scoped to the internal, single-estimator paths
measured in Section~\ref{sec:results-hessian}.

\subsection{Random copies of the trained models}\label{app:followups}

Section~\ref{sec:results-origin} of the main text asks whether the interaction
between labels comes with the architecture itself or appears only after
training, and reports that the normalized certificate $\bar X_{\mathrm{sep}}$
separates trained networks from untrained ones while $\anisoH$ does not
(Section~\ref{sec:results-training}). This subsection gives the construction and
the full range of values behind that comparison, together with a further
control that matches the entropy of the attention scores between trained and
random networks.

\paragraph{Random copies.} For each of the five models, we build $12$ random
copies and evaluate each copy on the same two datasets, \texttt{synthetic\_mi}
and \texttt{kin8nm}. Each copy is built by calling the reset method of every
module of the network, and we draw every parameter that these methods do not
reach from a Gaussian with standard deviation $0.02$. For \TabFM{}, we also
redraw from a standard normal the learned Fourier frequencies that its released
code stores outside the parameter list. The copy is restored only after the
standardization wrapper has been fit to the context, so that only the network's
own parameters differ from the trained model.

\paragraph{Results.} 
For the four models other than \TabFM{}, the trained values
are the means over the four queries of the first context of the grid behind
Table~\ref{tab:crossx}, and each random copy is evaluated on the same context
rows and labels with the first two of these queries. On \texttt{kin8nm}, the
feature scaling of the two inputs also differs slightly, because it is fit on
the context and query rows together. For \TabFM{}, the
trained values come instead from the first context and the first two queries of
that grid, which match the shared context exactly on \texttt{synthetic\_mi} and
differ from it on \texttt{kin8nm} only by a feature rescaling of at most $0.02$.
This comparison separates the training-associated share of the curvature from
the generic curvature that any nonlinear network carries even before training.
Every query of a trained model gives $\bar X_{\mathrm{sep}}\ge29.5\%$, with
means of $44\%$ to $98\%$, whereas the random copies stay at most $1.7\%$ as a
two-query mean and at most $3.4\%$ for a single query.
The gradient-scaled certificate
$\crossx/\lVert\PT a\rVert$ shows the same separation: $2.49$ to $21.7$ for the
trained models against a maximum of $0.037$ to $0.32$ across the random copies.
Random copies still carry generic curvature, with a mean
$\anisoH/\lVert\PT a\rVert$ of $2.0$ to $2.3$, so the corrected curvature norm
$\anisoH$ is nonzero for trained and random networks alike and, as a yes-or-no
certificate, cannot distinguish them. For the four models
other than \TabFM{}, the naive permutation
statistic $\permdelta$ of the trained models falls within the range of $12$
separately drawn random copies on all eight settings (at $n=128$ for all models
but \TabICL{}), so this simpler statistic cannot separate trained from
random either, in line with the same limit noted for $\anisoH$ in the main
text.

\paragraph{Attention-entropy control.} A further control asks whether matching
the entropy of the attention scores, rather than leaving random copies at their
default entropy, closes the gap. It covers three random copies per architecture
on two architectures (\TabICL{} and \TabPFNvtwosix{}) and two datasets, for $12$
evaluations. Matching the attention entropy between the random copies and the
trained network raises $\bar X_{\mathrm{sep}}$ only to $0.16$ for \TabICL{} and
to $0.004$ for \TabPFNvtwosix{}, and raises the gradient-scaled certificate to
at most $1.33$, far short of the trained values above. The \TabDPT{} entropy comparison is descriptive rather than a matched
control, because scaling its logits did not match its entropy to the trained
network's. Taken together, these controls support a training-associated change
in how labels interact, without isolating that change as the only one that
training introduces. Appendix~\ref{app:trajectory} follows the label response
of small transformers through training, in models whose architecture we control.

\subsection{Paths through the attention scores}\label{app:mechanism}

Attention combines the representations with weights that a softmax computes
from the attention scores. Section~\ref{sec:results-routing} shows that blocking
derivatives through them removes most of the measured interaction between
labels while leaving the prediction unchanged (Figure~\ref{fig:jointattention}b),
and Table~\ref{tab:main} reports the resulting fall in $\anisoH$ and $\crossx$
for the exact Hessian. This subsection reports that comparison in full, gives a
direct comparison between the intervened and the full curvature operator, states
the scope of this derivative-path intervention, and then reports a second
intervention that changes the forward computation itself.

We apply the intervention at every attention step the network executes: the
attention scores keep their forward values unchanged, so the prediction does not
change, while derivatives with respect to the labels are blocked from passing
through them. Writing the resulting Hessian as $H_{\mathrm{det}}$, the
difference $H-H_{\mathrm{det}}$ isolates the curvature carried by the blocked
derivative paths. For \TabICL{}, one label vector feeds both the column-wise
set transformer and the label encoder of the in-context stage
(Appendix~\ref{app:arms}), and a replay of the full pipeline lets the
derivatives pass through both sites, with forward-agreement checks confirming
that the replay leaves the prediction unchanged. \TabPFNiii{} and \TabFM{}
likewise receive one label vector at both of their label sites, and we
differentiate their full forward pass, so their derivatives also pass through
both sites.

\begin{table}[t]
\centering\small
\caption{\textbf{Attention-path norm reductions.} Every entry is the relative
fall of a norm under the intervention described in the text, not a share of
squared curvature and not a loss of accuracy. Values average over four datasets,
with three contexts per dataset and four queries per context (eight for the probe
estimate of $\anisoH$) at $n=256$ context rows. The
probe columns estimate the tangent curvature norm $\lVert\PT H\PT\rVert_F$ from
$16$ shared probes and $\anisoH$ from $64$ probes, and the exact columns use
Hessians assembled column by column on all $20$ settings. We measured \TabFM{}
with the exact Hessian only, so the probe means cover the other four models.}
\label{tab:routingconsolidated}
\begin{tabular}{lcccc}
\toprule
& \multicolumn{2}{c}{Probe estimate} & \multicolumn{2}{c}{Exact Hessian} \\
\cmidrule(lr){2-3}\cmidrule(lr){4-5}
Model & $\lVert\PT H\PT\rVert_F$ & $\anisoH$ & $\anisoH$ & $\crossx$ \\
\midrule
\TabICL{}    & $82.1\%$ & $82.8\%$ & $81.6\%$ & $82.0\%$ \\
\TabPFNiii{} & $87.7\%$ & $88.6\%$ & $85.9\%$ & $90.0\%$ \\
\TabPFNvtwosix{}      & $91.5\%$ & $91.4\%$ & $90.9\%$ & $95.3\%$ \\
\TabFM{}      & -- & -- & $78.2\%$ & $82.1\%$ \\
\TabDPT{}     & $69.6\%$ & $70.6\%$ & $68.0\%$ & $75.3\%$ \\
\midrule
Mean               & $82.7\%$ & $83.4\%$ & $80.9\%$ & $85.0\%$ \\
\bottomrule
\end{tabular}
\end{table}

A direct comparison of the full and the intervened curvature operator supports
the same reading. The part of the curvature attributable to the blocked
derivative paths, $\lVert \PT(H-H_{\mathrm{det}})\PT\rVert_F/\lVert \PT H\PT\rVert_F$, reaches
$0.96$, $0.99$, $1.00$, $0.94$, and $0.93$ for \TabICL{}, \TabPFNiii{},
\TabPFNvtwosix{}, \TabFM{}, and \TabDPT{}, in that order, while the part that
survives the intervention, $\lVert \PT H_{\mathrm{det}}\PT\rVert_F/\lVert \PT H\PT\rVert_F$, is
$0.19$, $0.14$, $0.09$, $0.22$, and $0.32$. Averaged over the twelve Hessians of
each setting, the cosine between the two parts lies between $-0.12$ and $0.28$,
consistent with a small remainder and limited
cancellation between them.

This decomposition is relative to the evaluated computational graph, and an
algebraically equivalent implementation could redistribute the derivative paths
between the two terms. The intervention also only blocks derivatives, so the
forward computation and the prediction stay unchanged. It therefore does not by
itself validate a model in which the attention scores are frozen at their
base-label values and never recomputed as the labels change. Such a forward
intervention would cache every attention-score tensor at the base labels and
reuse it on every perturbed forward pass, and measuring its own curvature would
require that frozen map's own gradient in the forward second-difference formula
of Eq.~\eqref{eq:intrinsic-directional-appendix}, together with a comparison
against that map's own prediction differences. Recomputing the attention scores
with a stop-gradient supplies neither. The ambient quotient kept in the
supplementary records omits these gradient terms as well, so it does not
identify a signed share of the intrinsic curvature either. The norm reductions
reported above therefore keep their scope as comparisons between derivative
paths in the computational graph as evaluated.

\paragraph{Replacing the cached attention maps.}\label{app:route}
We report a second intervention that does change the forward computation, as a
stress test of how much the attention scores contribute to the prediction
itself, rather than only to its derivatives. We run two forward passes. The
first pass computes the network on one set of labels and caches the attention
scores of every attention step in the order the network computes them. The
second pass recomputes the network's hidden states from a second, possibly
different set of labels, but at every attention step it replaces its own
attention scores with the corresponding cached ones, in the same order.

When both passes use the same labels, this hybrid computation reproduces the
native internal forward pass exactly on all $16$ settings of the four models
other than \TabFM{}, which this intervention does not cover. Replacing the cached
scores by those computed from a zero label vector instead shifts the prediction
by $0.60$ to $0.97$ standard deviations of the model's own native predictions,
averaged within each model. Because the hybrid computation applies later-layer
cached scores to hidden states that differ from the ones on which those scores
were computed once the two passes receive different labels, the exact agreement
obtained under identical labels validates only that the replay reproduces the
native forward pass correctly. It does not by itself show that varying the
cached scores isolates a causal effect of the attention scores.

An affine map fitted from the replayed internal outputs to the package's own
predictions in the original label units reproduces these predictions to within
$0.96\%$ relative error. Replacing the cached
scores worsens the normalized mean squared error (NMSE, the mean squared error
divided by the variance of the query targets) on $13$ of the $16$ settings and
improves it on the remaining three, with a median ratio of $5.12$ between the
patched and the native error. Interpolating between the true-label scores and
the zero-label scores increases the displacement from the native prediction
monotonically on $15$ of the $16$ settings. Replacing the cached scores at a
single attention call can produce a larger displacement than replacing them at
every call, which shows cancellation across calls and excludes an explanation
in which each call's attention scores contribute additively to the total
displacement. Per-setting results are in the supplementary material.

\subsection{Controlled models}\label{app:trajectory}

The pretrained models evaluated above cannot be retrained under a
different wiring, so answering which components the measured interaction needs,
and whether attention scores that depend on the labels are needed at all for a
model to use labels well, requires models whose architecture we control
ourselves. This subsection trains small transformers in which we vary how labels
enter the network, whether LayerNorm is present, and whether the attention
scores are allowed to depend on the labels.

\paragraph{A controlled transformer on a Gaussian-process task.}
We train a three-layer, $64$-dimensional row-token encoder for $4000$ Adam
steps on a two-dimensional radial-basis-function Gaussian-process prior, under
either Gaussian observation noise or $15\%$ outlier contamination. Across three
seeds, the held-out loss falls from about $1.3$ to $0.02$ under Gaussian noise and
$0.03$ under contamination within $1000$ steps.
These models receive the raw labels without standardization, so every label
vector is a valid input and the ambient derivatives are determined by the
predictions, which makes the correction of Section~\ref{subsec:followon}
unnecessary here. We track two descriptors across training, computed directly on
these derivatives: the permutation response $\permdelta$
(Eq.~\ref{eq:permdelta}), which measures how much the sensitivity vector changes
under a permutation of the context labels, and the off-diagonal ratio
$\lVert H-\operatorname{diag}(H)\rVert_F/\lVert H\rVert_F$ of the ambient Hessian. The permutation response grows from
$\permdelta\approx0.35$ at initialization to $0.85\pm0.50$ under Gaussian noise
and $1.38\pm0.29$ under contamination, and the final off-diagonal ratio reaches
$0.44$ to $0.50$. Because the Gaussian task fixes the kernel and the noise level,
its Bayes-optimal predictor, the Gaussian-process posterior mean, is exactly
linear in the labels, so this learned nonlinearity alone does not establish that
it is useful, and the higher response under contamination is suggestive but not
conclusive, given the spread across seeds. This small controlled model supports
no claim about scale.

\paragraph{LayerNorm is not necessary.}
Because LayerNorm's normalizing statistics are themselves computed from
representations that already mix the labels, we ask whether blocking
derivatives through the attention scores is confounded with blocking
derivatives through LayerNorm, or whether the two make separate contributions.
In a transformer of the same size trained on the contaminated task, blocking
derivatives through the
attention scores removes $92\pm2\%$ of the ambient off-diagonal norm across
three seeds, and blocking derivatives through both the attention scores and the
LayerNorm statistics removes $98\%$. Blocking the LayerNorm statistics alone
removes $85\%$, because normalization follows label mixing, so the two
interventions overlap rather than acting on separate parts of the interaction.
Yet a model trained with no LayerNorm at all reaches a comparable loss ($0.028$
against $0.026$ for the model with LayerNorm), keeps an off-diagonal ratio of
$0.41$, and still loses $97\%$ of its off-diagonal norm when derivatives cannot
pass through the attention scores. LayerNorm is therefore not necessary for the
interaction, since removing it leaves both the interaction and its
concentration in the attention scores in place.

\paragraph{Useful label processing without label-dependent attention scores.}
We next ask whether a model can process labels nonlinearly, and use that
nonlinearity effectively under contamination, without ever letting the
attention scores depend on the labels. A three-seed comparison matches depth,
width, number of heads, task, and training budget across four ways of wiring
labels into the network: injecting them into the same tokens from which the
attention scores are computed, as in four of the five pretrained models above
(token injection), injecting them only into the values that attention combines,
as in \TabDPT{} (value-only injection), computing the attention scores from the
features alone while a separate nonlinear stream processes the labels, so that
the labels cannot influence the attention scores at all (the score-blocked
model), and a fully label-linear control.

Under symmetric contamination, the score-blocked model reaches MSE
$0.0215\pm0.0008$, within $9\%$ of token injection's $0.0198\pm0.0009$, while
the label-linear control reaches $0.0388\pm0.0003$, $96\%$ worse than token
injection. Clean-task MSE ranges from $0.0145$ for token injection to $0.0214$ for
value-only injection. In the score-blocked
model, a smaller share of the curvature comes from interaction that no
one-label-at-a-time map can produce, $\bar{X}_{\mathrm{sep}}=0.085\pm0.145$
against $0.243\pm0.080$ for token injection.

A five-seed extension of this comparison covers $75$ runs with saved training
curves. Under symmetric contamination, the score-blocked and label-linear losses
are respectively $5\%$ and $91\%$ above token injection. Because the initial
score-blocked model has more parameters than token injection ($175{,}937$
against $100{,}865$), we also train a slim version with $107{,}993$ parameters,
$7\%$ more than token injection's count, which reaches MSE $0.0243\pm0.0016$:
within $20\%$ of token injection and $1.6$ times better than the label-linear
control. One-sided outlier contamination reproduces the same ordering, with the
score-blocked model $8\%$ above token injection against $142\%$ for the
label-linear control. The share of the curvature norm that passes through the
attention scores also changes with how labels enter the network, but its
direction reverses across contamination regimes: the share for value-only
injection is below token injection's under symmetric contamination ($0.829$
against $0.914$) and above it under one-sided contamination ($0.986$ against
$0.931$). These constructed models establish a useful alternative to
label-dependent attention scores, without identifying a necessary pathway in
the deployed models.

\subsection{How close a fixed-weight map comes}\label{app:tuning}

\begin{figure}[t]\centering
\includegraphics[width=\linewidth]{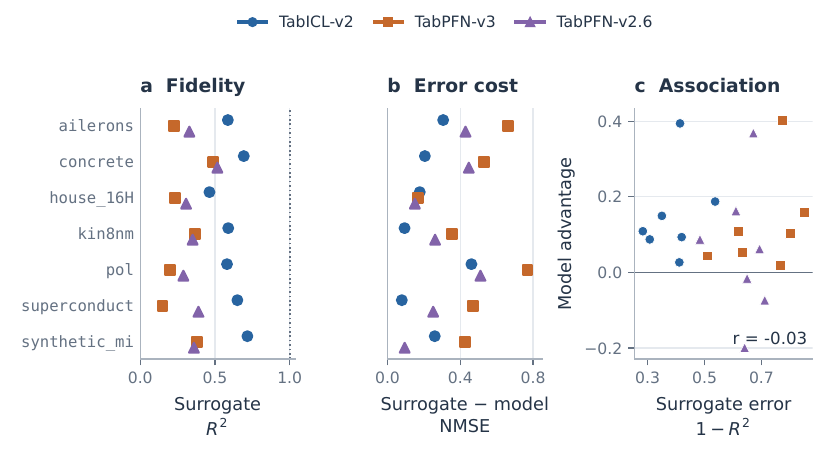}
\caption{\textbf{Approximation fidelity and predictive cost.} The same $21$
settings from three prior-fitted models show (a) the held-out
permutation-surrogate $R^2$, (b) the increase in NMSE when the model's own
predictions are replaced by this surrogate, and (c) the relation between the
surrogate's error and the model's predictive advantage, which is the held-out
NMSE of the best of three tuned label-linear smoothers ($k$-nearest neighbors,
kernel ridge regression and local-linear regression) minus the model's own NMSE,
so that a negative value means that a smoother predicts better than the model.
Replacement increases
the error on all $21$ settings, but the correlation between the two magnitudes
is $-0.03$, so this is an approximation comparison and not a causal ablation.
\TabDPT{} is not plotted here. A separate permutation fit with $200$ label draws
at $n=96$ gives it an $R^2$ of $-0.92$ and $-3.18$ on \texttt{synthetic\_mi} and
\texttt{kin8nm}, against $0.05$ and $0.03$ under the protocol of
Table~\ref{tab:taskpreserving}, so its permutation fidelity depends strongly on the
protocol.}
\label{fig:approximation}
\end{figure}

Section~\ref{sec:results-hessian} shows that no fixed-weight map reproduces any
of the five models exactly, since $\bar A_T$ is nonzero in every setting (Table~\ref{tab:main}).
Section~\ref{sec:results-positive} nonetheless reports that a fixed-weight
surrogate fitted to a model's predictions reaches a median held-out $R^2$ of
$0.72$, so failing the exact test does not mean that a model is far from such a
map on the labels it meets in practice. This subsection reports the comparators
behind that finding: an observed-feature comparator tuned on the context,
a surrogate fitted under a task-preserving label distribution, and a
constructed local-linear model with an explicit label-adaptive gain.
Figure~\ref{fig:approximation} relates the fidelity of a permutation-fitted
surrogate to the predictive advantage of the models.

\paragraph{An observed-feature comparator.}
A first comparator tunes Gaussian and dot-product kernel ridge regression,
$k$-nearest-neighbor regression, and local-linear regression on a held-out fifth
of the context rows, and we report the best of the four on the held-out
queries, a choice that favors the comparator. Each candidate map is
label-linear, while the tuning itself need not be. On $28$ settings, four models and seven
datasets at $n=96$ context rows, the selected comparator comes within $10\%$ of
the model's accuracy on $9$ of the $28$ settings, with a median held-out NMSE
gap of $+0.066$. A ridge surrogate fitted directly to the model's predictions
performs worse than this comparator on $27$ of the $28$ settings. This
comparator differs from the jointly trained, label-conditioned embeddings of
\citet{kernelicl}.

\paragraph{Task-preserving approximation.}\label{app:taskpreserving}
To keep the label distribution close to the one seen in practice rather than
far from it, we draw fresh observation noise for the synthetic task and
bootstrap the residuals of a gradient-boosted decision-tree (GBDT) fit on the
real datasets:
\[
y^{(k)}=\hat m(X)+\mathrm{resample}\bigl(y-\hat m(X)\bigr),
\]
where $\hat m$ is the GBDT fit to the data and $\mathrm{resample}(\cdot)$
redraws residuals with replacement. We fit raw-label surrogates on $210$ of $300$
label draws  and evaluate them on the other 
90, at $n=64$ context rows and six queries per setting.
Table~\ref{tab:taskpreserving} reports a median held-out $R^2$ of $0.723$ under
this task-preserving resampling, against $0.48$ under label permutations, while
the label-linear Nadaraya--Watson control is recovered exactly, at
$R^2=1.0000$ on all $12$ settings.

\begin{table}[t]\centering\small
\caption{\textbf{Held-out surrogate fidelity.} Held-out $R^2$, averaged over the six queries, of a
surrogate that is affine in $\yctx$, fitted to each model's predictions under
three label-draw distributions: $R^2_{\text{task}}$ under the task-preserving
resampling described in the text, $R^2_{\text{perm}}$ under label
permutations, and $R^2_{\text{NW-lin}}$ when the same fitting protocol is
applied to a label-linear Nadaraya--Watson control instead of a model. A value
of $1$ means that the surrogate reproduces the predictions exactly on held-out
draws, so the control column shows that the protocol recovers a genuinely
affine map whenever one is present. The bottom row gives the median over the
$12$ settings.}
\label{tab:taskpreserving}
\begin{tabular}{ll r r r}
\toprule
Model & Dataset & $R^2_{\text{task}}$ & $R^2_{\text{perm}}$ & $R^2_{\text{NW-lin}}$ \\
\midrule
\TabICL{} & \texttt{synthetic\_mi} & 0.97 & 0.68 & 1.00 \\
\TabICL{} & \texttt{kin8nm}        & 0.71 & 0.65 & 1.00 \\
\TabICL{} & \texttt{pol}           & 0.68 & 0.56 & 1.00 \\
\TabPFNiii{}  & \texttt{synthetic\_mi} & 0.93 & 0.28 & 1.00 \\
\TabPFNiii{}  & \texttt{kin8nm}        & 0.75 & 0.55 & 1.00 \\
\TabPFNiii{}  & \texttt{pol}           & 0.57 & 0.13 & 1.00 \\
\TabPFNvtwosix{} & \texttt{synthetic\_mi} & 0.99 & 0.42 & 1.00 \\
\TabPFNvtwosix{} & \texttt{kin8nm}        & 0.69 & 0.53 & 1.00 \\
\TabPFNvtwosix{} & \texttt{pol}           & 0.74 & 0.28 & 1.00 \\
\TabDPT{}    & \texttt{synthetic\_mi} & 0.95 & 0.05 & 1.00 \\
\TabDPT{}    & \texttt{kin8nm}        & 0.57 & 0.03 & 1.00 \\
\TabDPT{}    & \texttt{pol}           & 0.71 & 0.62 & 1.00 \\
\midrule
\multicolumn{2}{l}{median} & \textbf{0.72} & 0.48 & 1.00 \\
\bottomrule
\end{tabular}
\end{table}

Residual resampling of this kind removes the feature-dependent structure of the
noise. A separate check on \texttt{kin8nm}, \texttt{pol} and \texttt{superconduct} instead uses wild-bootstrap
multipliers, Rademacher and Mammen, which rescale each row's own residual by a
random factor, of magnitude one for the Rademacher multiplier. The median $R^2$ is $0.708$ under i.i.d.\ residual resampling,
$0.752$ under the Rademacher multiplier, and $0.771$ under the Mammen
multiplier, with a median span of $0.097$ across the three within each setting.
These resampling schemes are proxies for the task's label distribution,
distinct from the fixed-moment manifold $\mathcal M$ that the certificates of
Section~\ref{subsec:followon} use. Because permutations may lie farther from
\TabDPT{}'s real-data pretraining distribution, comparing near-task fidelity
across models is the safer choice here than comparing permutation fidelity. A
local Taylor-remainder check failed numerical stability on $2$ of the $12$
settings, so it supplies no claim about the remainder curve.

\paragraph{A constructed local model.}\label{app:positive}
A third comparator asks whether a simple, interpretable label-adaptive map can
reproduce the models' behavior, rather than only matching their predictions. It
weights each context row by a Gaussian kernel in the features together with an
exponential gain in that row's own deviation from the mean context label,
before a local-linear readout that uses these weights:
\[
\tilde w_i=\exp\!\left(-\frac{\lVert \xstar-x_i\rVert^2}{2h^2}\right)\exp\bigl(\lambda(y_i-\bar y)\bigr),
\]
where $h$ is the kernel bandwidth and $\lambda$ sets the strength of the
label-adaptive gain. Choosing $(h,\lambda)$ reproduces the aggregate response
profile on three synthetic tasks with $3$, $8$, and $16$ indices at $n=96$, and setting $\lambda=0$ removes the
label adaptation while still allowing the local-linear readout to carry signed
weights.

This construction nonetheless fits the trained models poorly. On the nine
settings of \TabPFNiii{}, \TabPFNvtwosix{} and \TabDPT{} on \texttt{synthetic\_mi},
\texttt{kin8nm} and \texttt{pol} at $n=128$, its median held-out
$R^2$ is $0.02$, against approximately one for maps that already belong to this
class. A variant that repeats the reweighting, recomputing the gain of each row
from its deviation from the previous prediction for up to eight rounds, does not
fit better, since its fit selects a single round on $8$ of the $9$ settings.

\end{document}